\documentclass[journal]{IEEEtran}

\ifCLASSINFOpdf
 
\fi

\usepackage[numbers,sort&compress]{natbib} %
\usepackage{tabularx}
\usepackage{subfigure}
\usepackage{makecell}
\usepackage[flushleft]{threeparttable}
\usepackage{algorithmic}
\usepackage{amsmath}
\usepackage{graphicx}
\graphicspath{{Figures/}}

\usepackage{rotating}
\usepackage{multirow}
\usepackage[ruled]{algorithm2e}
\usepackage{xcolor}
\def\HiLi{\leavevmode\rlap{\hbox to \hsize{\color{yellow!50}\leaders\hrule height .8\baselineskip depth .5ex\hfill}}}

\begin{document}
%
\title{Knowledge Transfer for Dynamic Multi-objective Optimization with a Changing Number of Objectives}

\author{Gan~Ruan,~Leandro~L.~Minku, \textit{Senior Member, IEEE,}~Stefan~Menzel,~Bernhard~Sendhoff,~\textit{Fellow, IEEE,} and~Xin~Yao, \textit{Fellow, IEEE}
\thanks{Gan Ruan, Leandro L.Minku and Xin Yao are with CERCIA, School of Computer Science, University of Birmingham, Edgbaston Birmingham B15 2TT, UK (e-mail: GXR847@student.bham.ac.uk, L.L.Minku@bham.ac.uk, xiny@sustech.edu.cn ).}
\thanks{Stefan Menzel and Bernhard Sendhoff are with the Honda Research Institute Europe GmbH, 63073 Offenbach, Germany. (email: stefan.menzel@honda-ri.de, bernhard.sendhoff@honda-ri.de)}
\thanks{Xin Yao is also with the Department of Computer Science and Engineering, Southern University of Science and Technology, Shenzhen, China.}
}

\maketitle

\begin{abstract}
Different from most other dynamic multi-objective optimization problems (DMOPs), DMOPs with a changing number of objectives usually result in expansion or contraction of the Pareto front or Pareto set manifold.  Knowledge transfer has been used for solving DMOPs, since it can transfer useful information from solving one problem instance to solve another related problem instance. However, we show that the state-of-the-art transfer algorithm for DMOPs with a changing number of objectives lacks sufficient diversity when the fitness landscape and Pareto front shape present nonseparability, deceptiveness or other challenging features. Therefore, we propose a knowledge transfer dynamic multi-objective evolutionary algorithm (KTDMOEA) to enhance population diversity after changes by expanding/contracting the Pareto set in response to an increase/decrease in the number of objectives. This enables a solution set with good convergence and diversity to be obtained after optimization. Comprehensive studies using 13 DMOP benchmarks with a changing number of objectives demonstrate that our proposed KTDMOEA is successful in enhancing population diversity compared to state-of-the-art algorithms, improving optimization especially in fast changing environments.
\end{abstract}


%
\IEEEpeerreviewmaketitle

\section{Introduction}
\label{sec:Intro}
Dynamic multi-objective optimization problems (DMOPs) \cite{farina2004dynamic}, widely existing in the real-world \cite{application_scheduling_deb2007dynamic}, are a kind of multi-objective optimization problems which comprise a series of problems whose objectives change over time \cite{ruan2017effect}. Due to the dynamics in objective functions of DMOPs, the Pareto sets (PSs) and/or Pareto fronts (PFs) may change over time. Therefore, how to efficiently track the changing PSs/PFs is a key problem in solving DMOPs. Facing this challenge, many strategies have been proposed to tackle the dynamics in DMOPs. They can be classified as diversity enhancement \cite{goh2009competitive}, \cite{peng2014population}, diversity introduction \cite{application_scheduling_deb2007dynamic}, \cite{zheng2007new}, \cite{liu2010sphere}, memory techniques \cite{2019Hybrid}, \cite{koo2010predictive}, prediction strategies \cite{hatzakis2006dynamic, zhou2014population, muruganantham2013dmoea, muruganantham2015evolutionary} and knowledge transfer-based methods \cite{jiang2017transfer,  jiang2020individual, jiang2020fast, jiang2020knee}.



However, few studies have been done to solve DMOPs with a changing number of objectives (NObj). The most recent work is the proposal of the Dynamic Two Archive Evolutionary Algorithm (DTAEA) \cite{chen2017dynamic}. The main idea was to simultaneously maintain two co-evolving populations, i.e, a convergence archive (CA) and a diversity archive (DA) during the evolution. Whenever environmental changes occur, CA and DA are reconstructed to preserve as much convergence and diversity as they can in the new environment.

Considering that the reconstruction of CA and DA in DTAEA involves copying (optimal) solutions from the past problem instance to the next after changes, DTAEA can be regarded as a kind of knowledge transfer-based algorithm, as it makes use of knowledge acquired from solving the previous problem instance to solve the new problem instance. However, in this study, we show that DTAEA cannot handle DMOPs with a changing NObj containing more complex problem features including PF shapes (convex, discontinuous and mixed shape of convex and concave) and fitness landscapes (nonseparability and deceptiveness) well. Specifically, the knowledge transfer (i.e. CA and DA reconstruction) in DTAEA is incapable of providing enough diversity in these complex scenarios. The reason is that the change in the NObj changes the distribution of reconstructed CA on the true PF for the new problem instance and the uniformly sampled solutions in the search space by the DA reconstruction are not uniformly distributed in the objective space due to problem features in the more complex problems.

In this paper, we aim to answer the following research questions:
\begin{itemize}
\item How to increase diversity when solving DMOPs with a changing NObj, so as to improve knowledge transfer right after changes?
\item How does knowledge transfer help the optimization process itself?
\end{itemize}

In order to answer these research questions, 
we propose to expand or contract the PS of the problem after NObj increases or decreases, respectively, to improve the knowledge transfer. This strategy works better than DTAEA because DMOPs with a changing NObj usually result in the expansion or contraction of the dimension of the PS manifold \cite{chen2017dynamic}. Experimental studies have been carried out on 13 DMOPs with a changing NObj, modified from 4 DTLZ \cite{deb2005scalable} and 9 WFG \cite{huband2006review} problems to demonstrate the effectiveness of our proposed approach.

The novel contributions of our work are summarized as follows:
\begin{itemize}
\item Comprehensive experiments have been carried out on representative problems with complex problem features in the fitness landscape (nonseparability and deceptiveness) and complex PF shapes (convex, discontinuous and mixed shape of convex and concave) to understand the limitations of the state-of-the-art algorithm DTAEA. Our analyses reveal that DTAEA lacks diversity when solving more complex DMOPs with a changing NObj;
\item A novel knowledge transfer-based method, called knowledge transfer dynamic multi-objective evolutionary algorithm (KTDMOEA), is proposed. This method proposes PS expansion and contraction mechanisms to enhance diversity for dealing with changing NObj in DMOPs;
\item Systematic computational studies have been conducted to compare our proposed KTDMOEA with 5 algorithms on 13 DMOPs with a changing NObj under different frequencies and types of changes in the NObj. Experimental results have shown that our algorithm is competitive against all compared algorithms.
\end{itemize}

The remainder of this paper is organized as follows. Section \ref{sec:RelateWork} describes related work on DMOPs with a changing NObj and evolutionary transfer optimization as well as the motivation of our proposal. The proposed knowledge transfer-based algorithm is elaborated in Section \ref{sec:KTDMOEA}. Section \ref{sec:ExperSetup} describes the specific experimental setup. The experimental results are presented in detail in Section \ref{sec:results}. Section \ref{sec:conclusion} concludes this paper and points out possible future work.

\section{Related Work and Motivation}
\label{sec:RelateWork}

This section firstly reviews related work on DMOPs with a changing NObj and evolutionary transfer optimization. Then, a preliminary investigation of the existing work DTAEA \cite{chen2017dynamic} is conducted to reveal its limitations on solving DMOPs with complex problem features.

\subsection{DMOPs with a Changing NObj}

In this paper, we focus on the continuous minimized DMOPs defined as follows:
\begin{equation}\label{eq:def of dMOP}
\begin{split}
\begin{aligned}
\begin{cases}
\min~\mathbf{F}(\mathbf{x},t)=\big(f_1(\mathbf{x},t),\ldots,f_{m(t)}(\mathbf{x},t)\big)^T\\
s.t. ~~ \mathbf{x}\in \Omega,~ t\in \Omega_t
\end{cases}
\end{aligned}
\end{split}
\end{equation}
where $\Omega\subseteq R^n$ is the decision (variable) space; t is the discrete time instance; $\Omega_t \subseteq R$ is the time space. $\textbf{F}(\textbf{x},t): \Omega \times \Omega_t \to R^{m(t)}$ is the objective function vector that evaluates a candidate solution $\textbf{x}=(x_1,...,x_n)$ at time $t$. $m(t)$ is the number of objective at time $t$.


Although people have realized the importance of tackling DMOPs with a changing NObj and mention this concept in \cite{guan2005evolving, azzouz2014multiple, huang2011dynamic, jiang2019scalable}, few work existed studying this problem until recently \cite{chen2017dynamic}. Recently, a comprehensive investigation was conducted on the challenges of DMOPs with a changing NObj in \cite{chen2017dynamic}. It has been experimentally demonstrated that it is a key issue of how to propel crowded solutions to cover the whole PF and how to pull unconverged solutions back to the PF with good diversity when increasing and decreasing the NObj, respectively. Bearing this challenge in mind, the authors in \cite{chen2017dynamic} proposed DTAEA to tackle DMOPs with a changing NObj, in which two complementary populations, CA and DA, are simultaneously maintained in the evolution process to focus on population convergence and diversity, respectively. Whenever environmental changes occur, CA and DA are reconstructed to preserve as much convergence and diversity as they can in the new environment. More specifically, when increasing the NObj, solutions in the old CA are all copied to the new CA. When decreasing the NObj, nondominated and dominated solutions of the old CA are all copied to the new CA and new DA, respectively. Therefore, DTAEA can be seen as a kind of knowledge transfer-based algorithm, as it makes use of knowledge acquired from previous solutions.




\subsection{Evolutionary Transfer Optimization}

Knowledge transfer has been applied to evolutionary computation to solve mutli-objective optimization and dynamic multi-objective optimization problems \cite{tan2021evolutionary}. Specifically, knowledge transfer is able to learn useful knowledge from related problem instances to solve the targeted problem instance \cite{gupta2015multifactorial, gupta2017insights}. However, evolutionary multi-tasking optimization (EMT) \cite{gupta2015multifactorial, gupta2017insights, da2018curbing, feng2018evolutionary} differs from our scenario here because EMT considers solving multiple tasks simultaneously, while our work considers solving different problem instances sequentially as the environment, e.g., NObj, changes. At any given time, we solves only one problem instance, not multiple ones.



For dynamic multi-objective optimization, knowledge transfer can help to predict good solutions for the next change based on previously optimized solutions. The transfer learning-based dynamic multi-objective evolutionary algorithm (Tr-DMOEA) \cite{jiang2017transfer} was the first work of applying knowledge transfer to solve DMOPs, in which transfer component analysis is used to transfer solutions in the PF of the previous environment to generate an initial population for the next environment. An autoencoding evolutionary search is regarded as a knowledge transfer method to predict the moving of PSs based on nondominated solutions obtained before the change \cite{feng2020solving}. In \cite{jiang2020fast}, a manifold transfer learning method is applied to forecast the changing PSs over time. However, these knowledge transfer-based DMOEAs never considered changing NObj and cannot solve DMOPs with a changing NObj, since they were designed to track the changing position or shape of PSs or PFs rather than expand or contract PS/PF. In general, it is always very challenging to decide what, when and how to transfer in DMOPs \cite{ruan2019and, ruan2020computational}.

\subsection{A Preliminary Investigation of DTAEA}
\label{sec:PreDTAEA}



Even though DTAEA has been computationally demonstrated in \cite{chen2017dynamic} to be effective on DMOPs with a changing NObj based on knowledge transfer, the test problems that were used to evaluate DTAEA are somewhat limited, as problem features in those problems are relatively simple, such as linear or concave PF shape and fitness landscape with multimodality, bias or even nothing.


In order to evaluate whether DTAEA is able to solve DMOPs with a changing NObj and more complex problem features including PF shapes (convex, discontinuous and mixed shape of convex and concave) and fitness landscapes (nonseparability and deceptiveness), a benchmark problem WFG4 is arbitrarily selected from the WFG suite \cite{huband2006review} as an example to conduct an experimental investigation of the performance of DTAEA. In contrast, F2 \cite{chen2017dynamic} is arbitrarily selected from the DTLZ suite \cite{deb2005scalable}. When increasing the NObj, the problems are set as bi-objective problems and then given 1000 generations to evolve by DTAEA before increasing the NObj from 2 to 3. When decreasing the NObj, the problems are set as tri-objective problems and then given 1000 generations to evolve by DTAEA before decreasing the NObj from 3 to 2.

\begin{figure}
  \centering
  \subfigure[F2]{
    \label{fig:F2} 
    \includegraphics[width=0.48\linewidth]{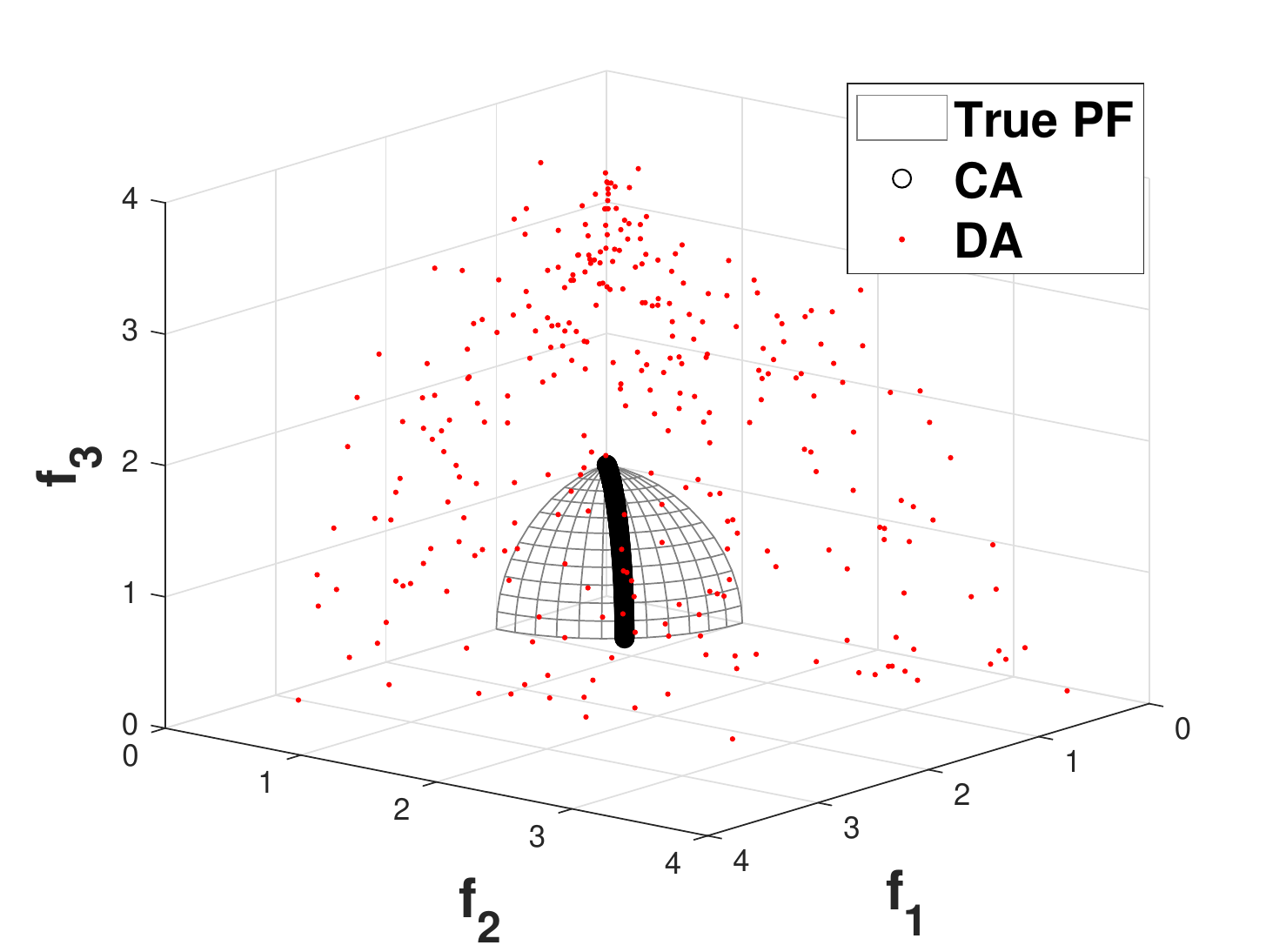}}
  \subfigure[WFG4]{
    \label{fig:WFG4} 
    \includegraphics[width=0.48\linewidth]{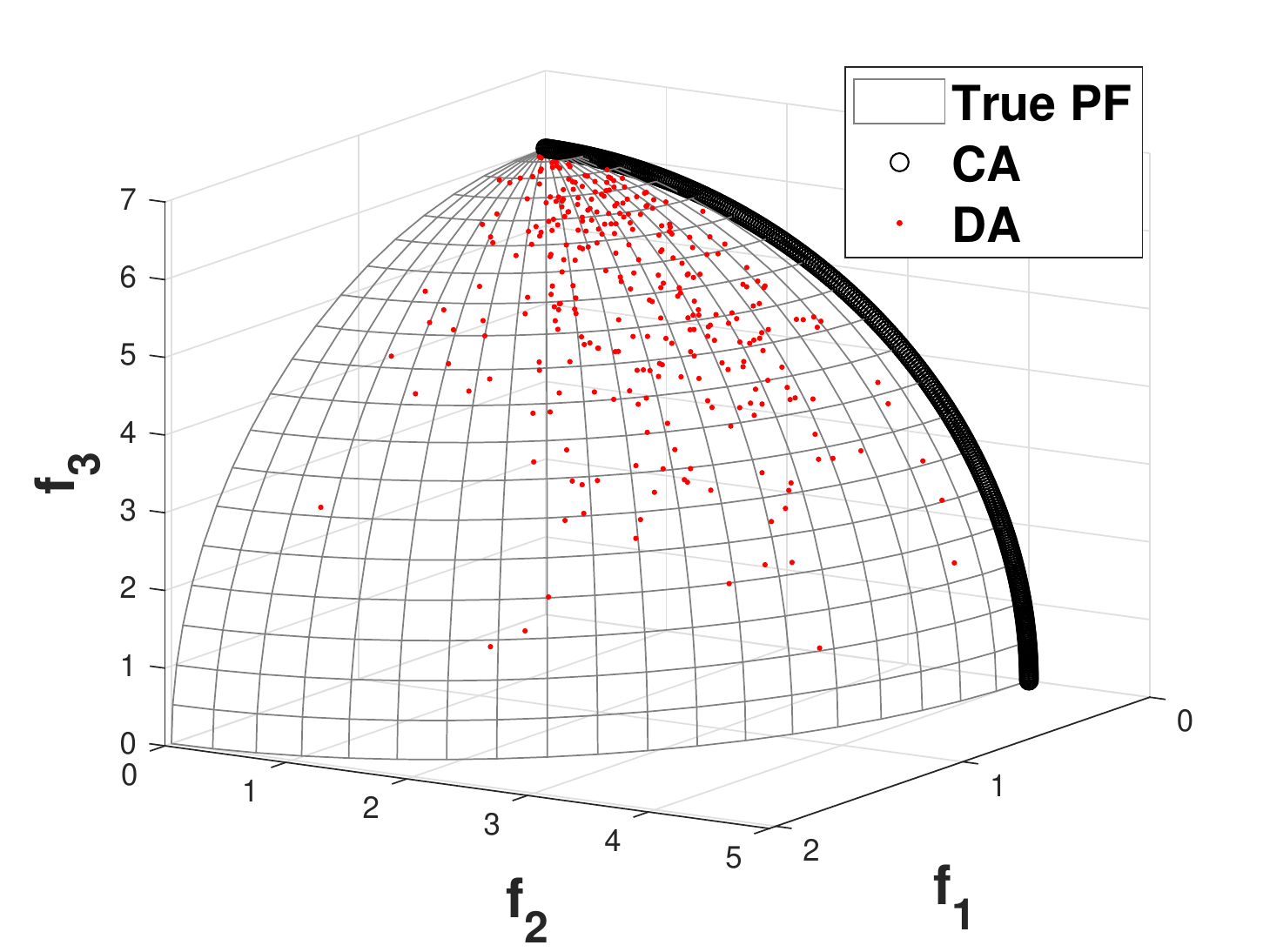}}\vspace{-0.35cm}
  \caption{Distribution of the reconstructed CA and DA obtained by DTAEA in the first generation right after changes when increasing NObj from 2 to 3 on F2 and WFG4.}
  \label{fig:increseF2WFG4} 
  \vspace{-0.4cm}
\end{figure}

\begin{figure}
  \centering
  \subfigure[F2]{
    \label{fig:F2} 
    \includegraphics[width=0.48\linewidth]{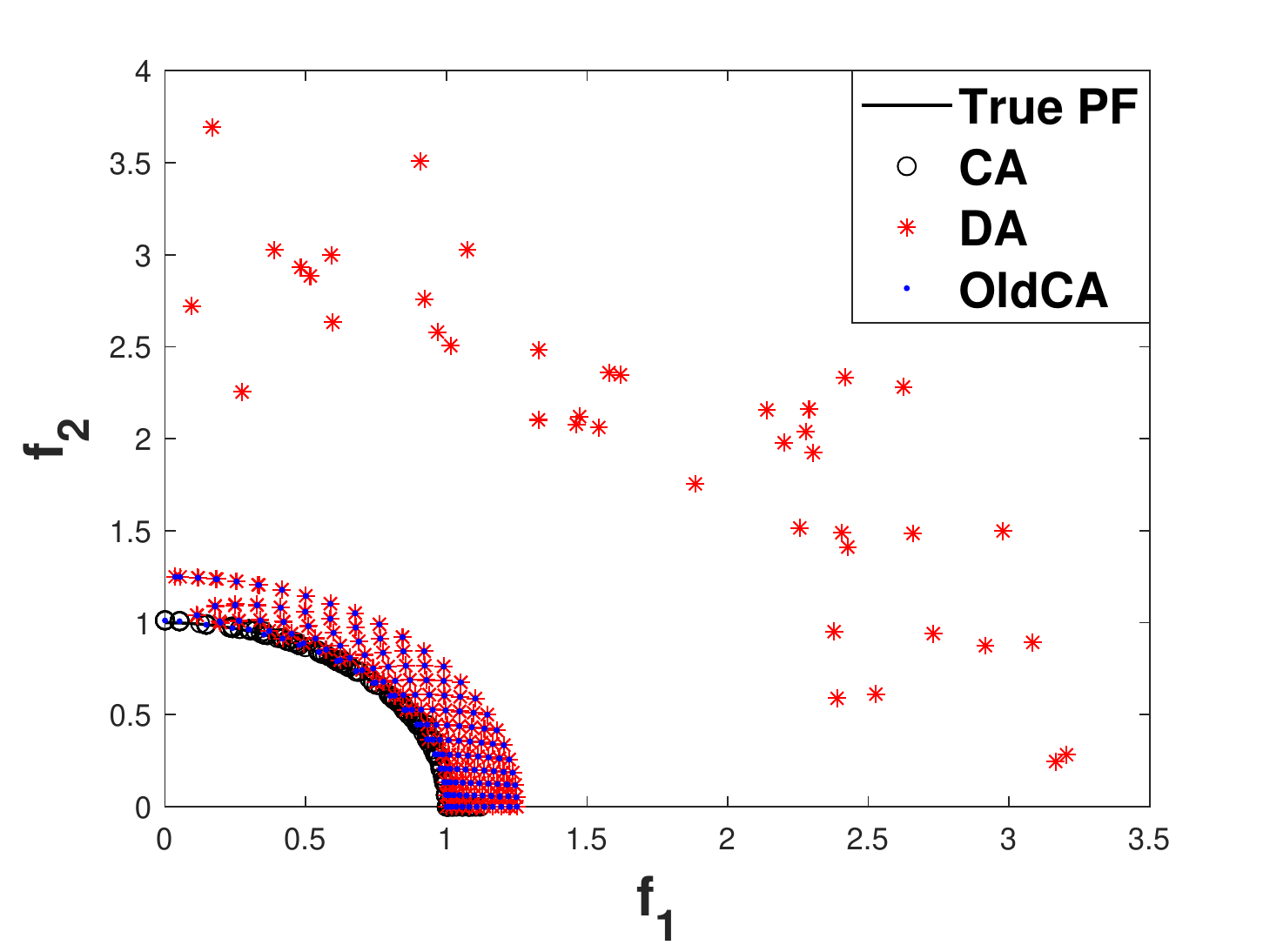}}
  \subfigure[WFG4]{
    \label{fig:WFG4} 
    \includegraphics[width=0.48\linewidth]{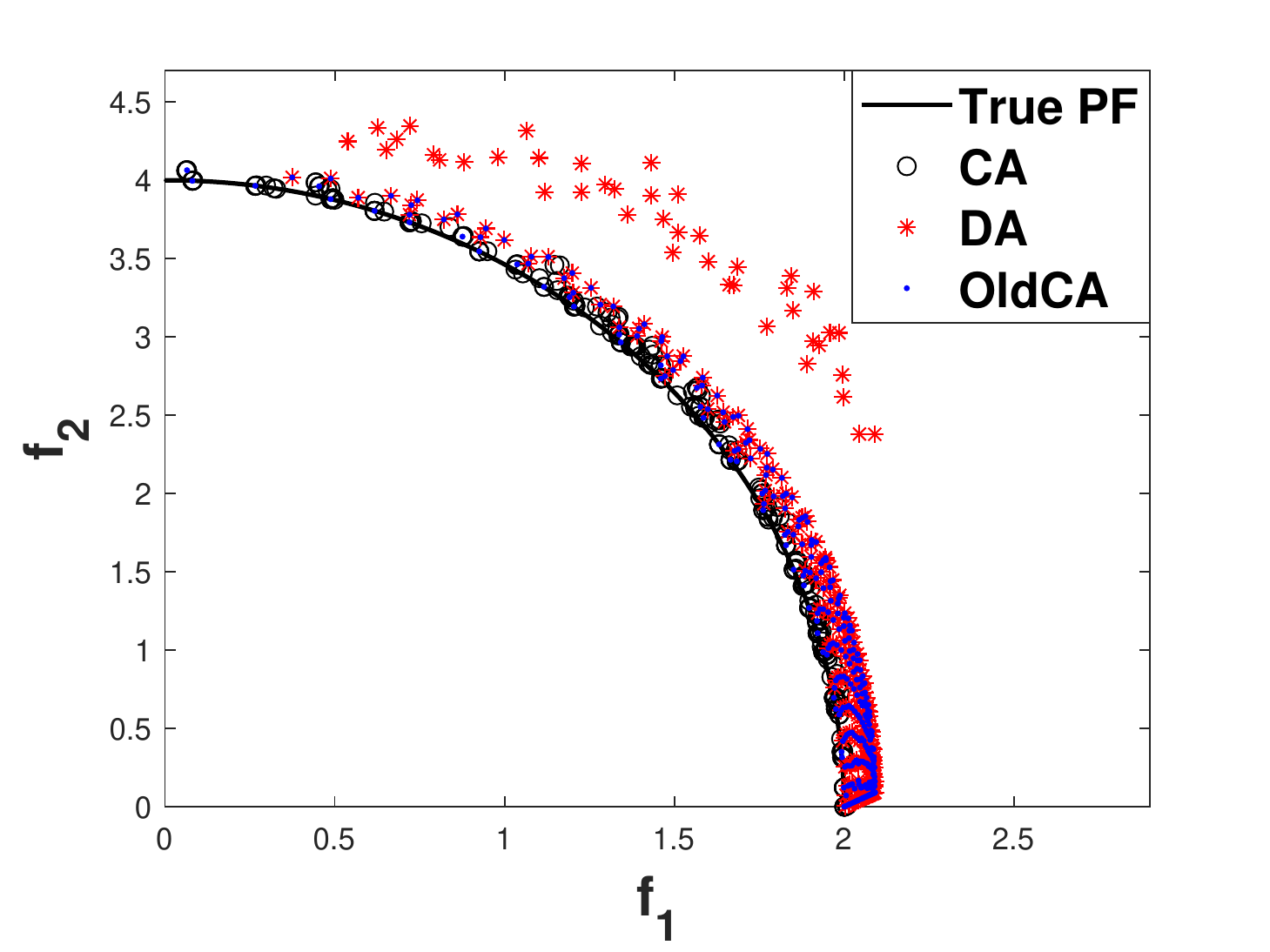}}\vspace{-0.35cm}
  \caption{Distribution of the reconstructed CA and DA obtained by DTAEA in the first generation right after changes when decreasing NObj from 3 to 2 on F2 and WFG4.}
  \label{fig:decreaseF2WFG4} 
  \vspace{-0.5cm}
\end{figure}

Figures \ref{fig:increseF2WFG4} and \ref{fig:decreaseF2WFG4} show the distribution of the old CA, the reconstructed CA and DA obtained by DTAEA on the two problems F2 and WFG4 in the first generation right after changes for the cases of increasing the NObj from 2 to 3 and decreasing the NObj from 3 to 2, respectively. It should be noted that when increasing the NObj, solutions in the old CA are all copied to the new CA. Therefore, in Figure \ref{fig:increseF2WFG4}, `CA' represents both solutions in the old CA and the new CA. It is clear from Figure \ref{fig:increseF2WFG4} that when increasing the NObj from 2 to 3, the new CA does not have good diversity on both F2 and WFG4. As for the reconstructed DA, it has a good level of diversity on F2. As shown in Figure \ref{fig:increseF2WFG4}, solutions randomly generated in the search space are covering the whole area over the true PF. However, on WFG4, solutions in the reconstructed DA only cover a part of the PF over it.

Similarly, it can be seen from Figure \ref{fig:decreaseF2WFG4} that solutions in DA also have good diversity on F2 when decreasing NObj from 3 to 2. However, on WFG4, there are some areas close to the high values of the second objective in the objective space without any solutions covered by the DA. It is clear that on both problems and the two cases (increasing and decreasing the NObj), the reconstructed CA and DA do not provide enough diversity. Therefore, the CA and DA reconstruction of DTAEA cannot provide enough diversity on DMOPs with more complex problem features. The reason is that the problem features in the more complex problems cause uniformly sampled solutions in the search space not to be uniformly distributed in the objective space.

\section{Knowledge Transfer Dynamic Multi-objective Evolutionary Algorithm (KTDMOEA)}
\label{sec:KTDMOEA}

\begin{figure}
\vspace{-0.8cm}
	\centering
	\includegraphics[width=0.5\textwidth]
	{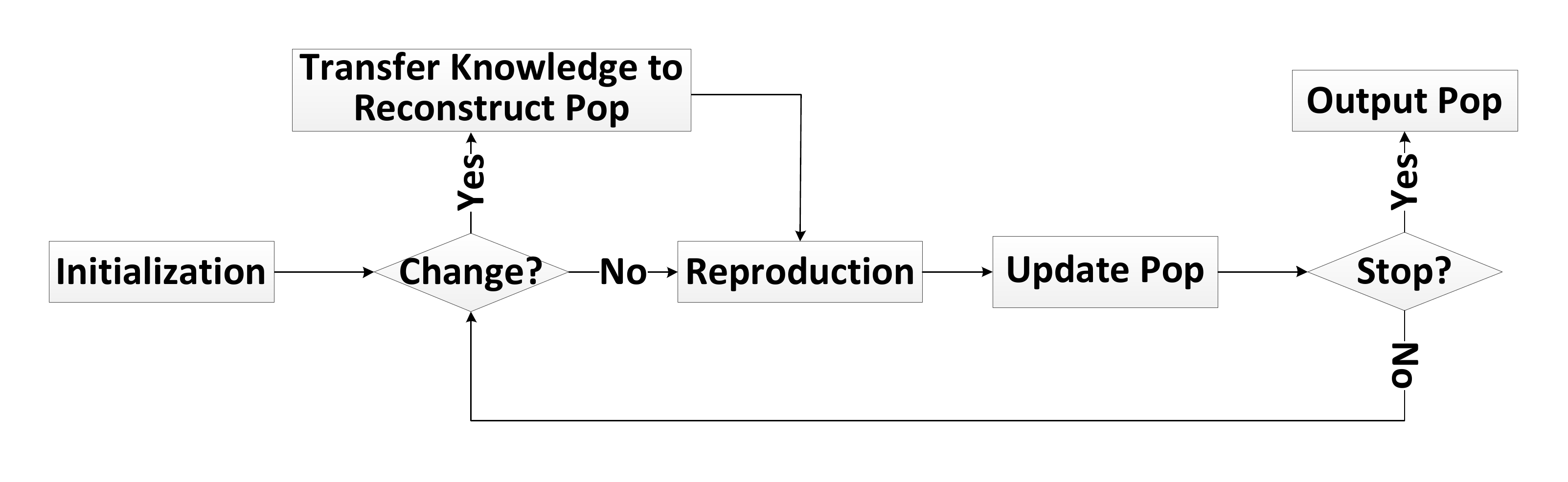}\vspace{-0.65cm}
	\caption{Flow chart of KTDMOEA.}
	\label{fig:flowKTMOEA}
	\vspace{-0.4cm}
\end{figure}

In this section, we present our proposed knowledge transfer dynamic multi-objective evolutionary algorithm (denoted as KTDMOEA), which is designed to tackle more complex DMOPs with a changing NObj. The main component of \mbox{KTDMOEA} is the proposed diversity enhanced knowledge transfer, which is designed to improve population diversity right after changes in DMOPs with a changing NObj. KTDMOEA's flowchart is shown in Figure \ref{fig:flowKTMOEA}. As the flowchart exhibits, KTDMOEA maintains a single population. Whenever there is a change in the NObj, the process of knowledge transfer is evoked to reconstruct the population such that it has increased diversity; otherwise, other procedures in the evolution process are carried out. The novel process of knowledge transfer through PS expansion/contraction proposed in this paper will be elaborated in Section \ref{sec:kt}. Then, Section \ref{sec:KTDMOEA_alg} presents the overall evolutionary process of the proposed KTDMOEA.

\subsection{Diversity Enhancing Knowledge Transfer}\label{sec:kt}
As DMOPs with increasing/decreasing NObj result in the expansion/contraction of PS/PF, we propose to enhance diversity through PS expansion and contraction for increasing and decreasing NObj, respectively. This strategy is targeted at  enhancing knowledge transfer right after changes. In this section, the specific details of PS expansion and contraction are given in Section \ref{ExpandPS} and Section \ref{ContractPS}, respectively.

\subsubsection{Expand the PS when Increasing the NObj}
\label{ExpandPS}

Increasing the NObj usually results in the expansion of the dimension of PS/PF manifold. Therefore, the PS is proposed to be expanded when increasing the NObj, so as to increase the population diversity. 

\begin{figure}
  \centering
  \subfigure[PS expansion]{
    \label{fig:PSExpand} 
    \includegraphics[width=0.48\linewidth]{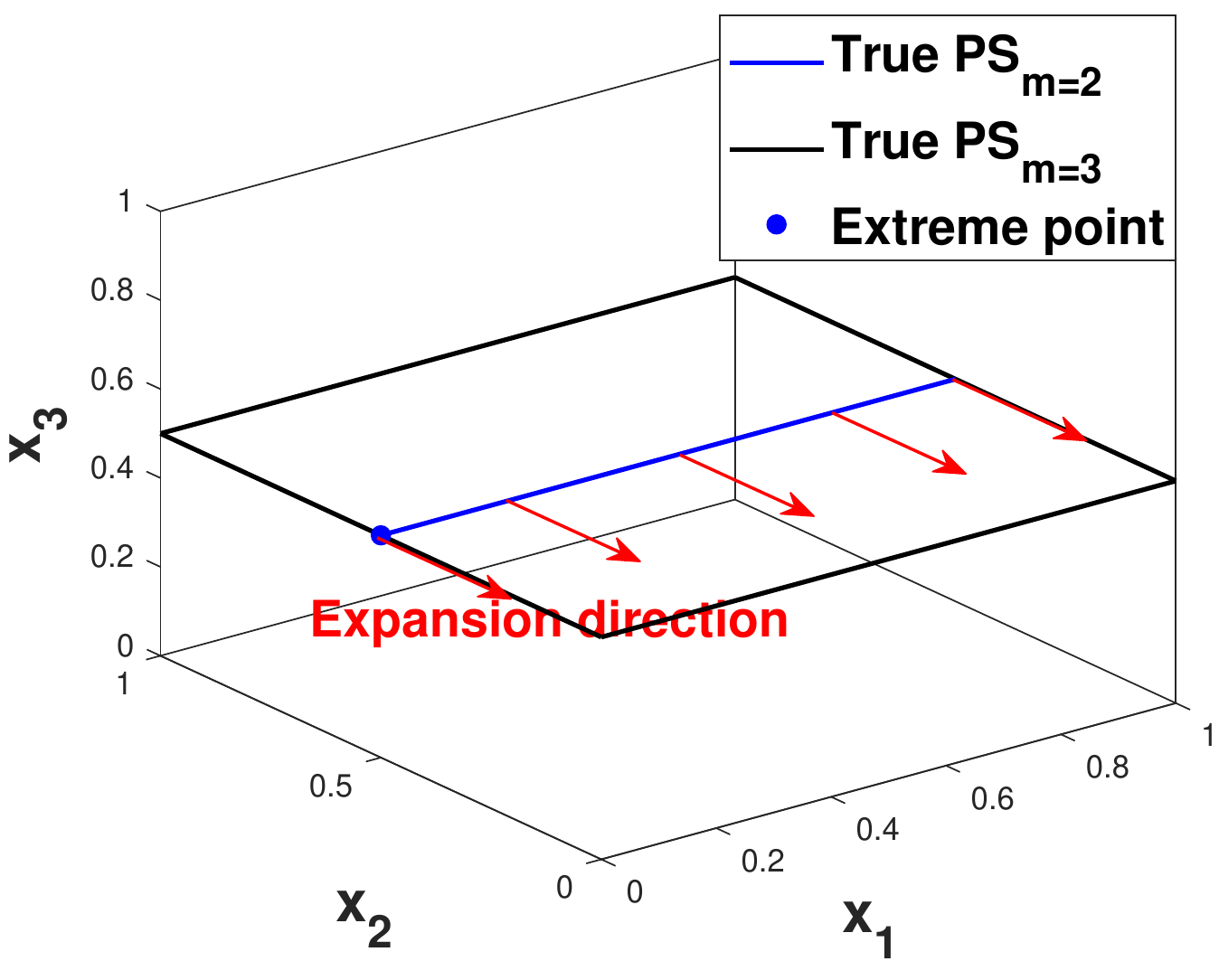}}
  \subfigure[PS contraction]{
    \label{fig:PSContract} 
    \includegraphics[width=0.48\linewidth]{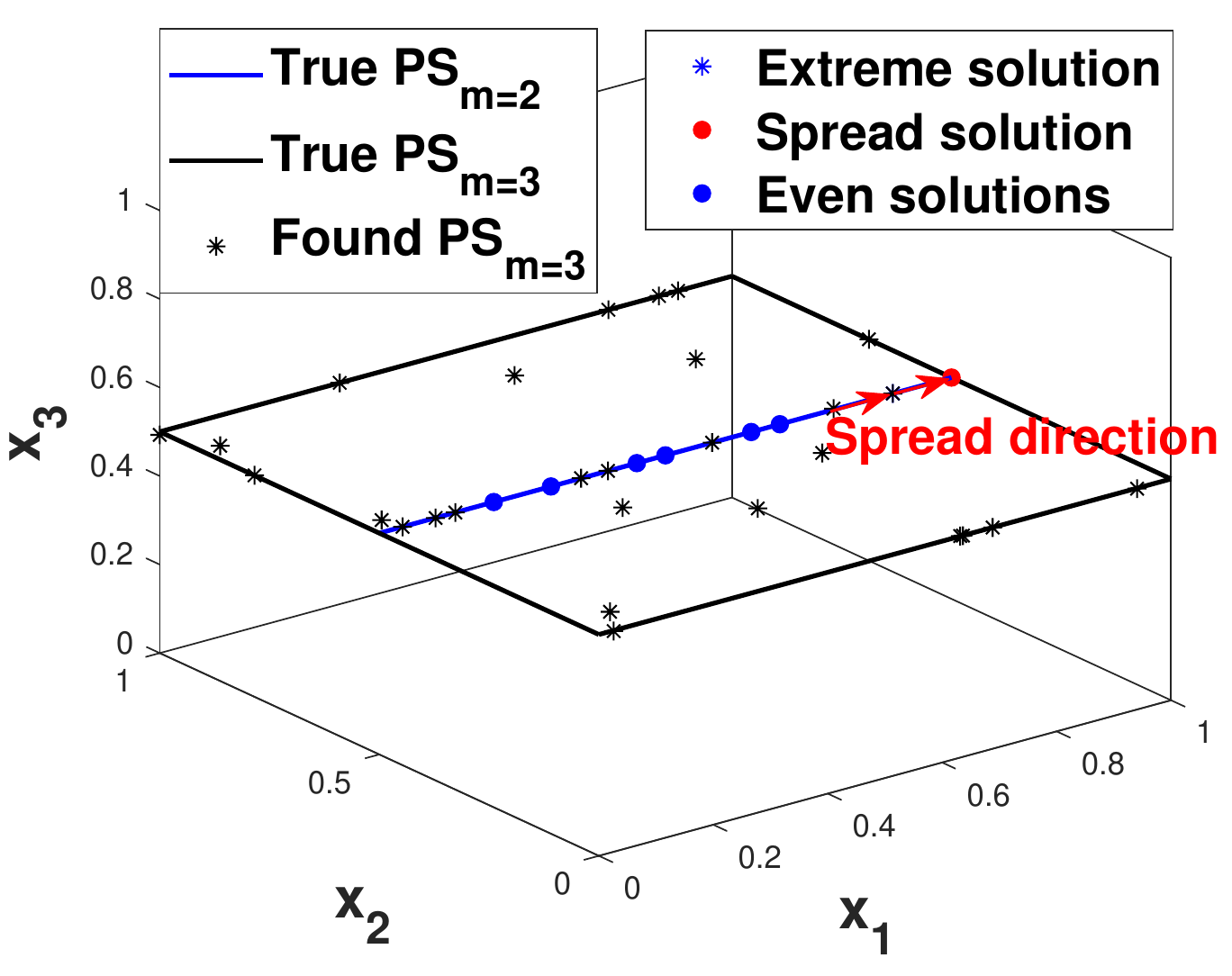}}\vspace{-0.2cm}
  \caption{Brief illustration of how to expand and contract the PS for increasing and decreasing NObj, respectively.}
  \label{fig:KT} 
  \vspace{-0.4cm}
\end{figure}

The idea of PS expansion in the decision space is illustrated in Figure \ref{fig:PSExpand}. Note this figure is just drawn to demonstrate the process of PS expansion, the specific PF and expansion direction in real problems may be different. As shown in the figure, suppose the blue point is one extreme point in the PS before the change ($PS_t$); blue line is the Pareto optimal set at time step t with two NObj; the expansion direction is found by generating several solutions around the blue point and connecting the blue point to the point nondominated to it; the plane formed by 4 black lines is the Pareto optimal set at time step $t+1$ with three NObj and the red arrows are the expansion directions. Solutions evenly selected in $PS_t$, which are the points in the starting points of the red arrows, are regarded as the PS expansion base solutions to cover the whole PS right after the change ($PS_{t+1}$). 

\begin{algorithm}[ht!]
\KwIn{Pareto optimal solution set at time t ($\textbf{PS}_t$); set of the searched expansion directions $\textbf{D}$ with size $N_{dir}$; population size $N$; the NObj for the old problem before the change $M_{t}$; number of solutions to generate along each expansion direction $\theta$;}
\KwOut{Transferred solutions $\textbf{P}_{tr}$}
\nl Search a set of expansion direction as $\textbf{D}$ via Algorithm \ref{Alg:SearchExpanDir}\;
\nl Evenly select solutions from $\textbf{PS}_t$ in the objective space with the size of $N_{base} = \lfloor \frac{N - M_{t}}{N_{dir} * \theta} \rfloor$ as $\textbf{P}_{base}$\;
  \nl Generate $\theta$ solutions along each direction in $\textbf{D}$ to fill $\textbf{P}_{tr}$ through the following equation \ref{eq:expand}\;
  
  \nl Evenly select solutions from $\textbf{PS}_t$ in the objective space with the size of $N - N_{base} * \theta * N_{dir}$ to fill $\textbf{P}_{tr}$\;
 \textbf{Return} $\textbf{P}_{tr}$.
    \caption{{\bf Expand the PS to generate transferred solutions} \label{Alg:PSExpansion}}
\end{algorithm}

The framework of PS expansion is exhibited in Algorithm \ref{Alg:PSExpansion}. As Algorithm \ref{Alg:PSExpansion} presents, in order to achieve PS expansion, the first step is to search for the potential PS expansion directions, whose procedure is given in Algorithm \ref{Alg:SearchExpanDir} and explained as follows. Given a set of Pareto optimal solutions at the time step t ($\textbf{PS}_t$), the algorithm firstly finds solutions with the maximum objective value for each objective as the set of extreme points (denoted as $\textbf{P}_e$) in line 1 of Algorithm \ref{Alg:SearchExpanDir}. The use of extreme points helps to ensure that the found expansion directions are not misleading. If the middle points in the PF are selected, the found expansion directions might be along the direction of the PF, thus wasting computational resources. Then, in line 2, a random extreme point $\textbf{x}_e$ is selected from the set $\textbf{P}_e$ as the initial point of expansion directions. Later on, a solution set $\textbf{P}_{var}$ with the same size as the population size is randomly produced around $\textbf{x}_e$ via the polynomial mutation \cite{deb1995simulated} to be regarded as the candidate sets of end points of expansion directions in line 3, which is also called detective population.

\begin{algorithm}[ht!]
\KwIn{Pareto optimal solution set at time t ($\textbf{PS}_t$)}
\KwOut{Set of the searched expansion directions $\textbf{D}$ or NULL.}
\nl Find the solutions with the maximum objective value for each objective as the set of extreme points ($\textbf{P}_e$)\;
  \nl Randomly select one solution from $P_e$ and regard it as $\textbf{x}_e$\;
  \nl Produce a solution set $\textbf{P}_{var}$ (also called detective population) with the same size as the population size through randomly generating solutions via the polynomial mutation \cite{deb1995simulated} around the selected extreme point $x_e$\;
  \nl Evaluate all solutions in $\textbf{P}_{var}$ and delete dominated solutions after conducting nondominated sorting on them\;
  \nl \If{All solutions in $\textbf{P}_{var}$ are nondominated by $\textbf{PS}_t$}{
  \nl Set $\textbf{P}_{non}$ as $\textbf{P}_{var}$\;}
  \nl \Else
  {
  \nl Delete solutions from $\textbf{P}_{var}$ that are dominated by those in the $\textbf{PS}_t$ and set the remaining solutions as $\textbf{P}_{non}$\;
      }
   \nl Use evenly generated weight vectors following the method in \cite{chen2017dynamic} to estimate density of $\textbf{P}_{non}$ and $\textbf{PS}_t$ with the method introduced in Section \ref{sec:KTDMOEA_alg}\;
   \nl Delete solutions in $\textbf{P}_{non}$ located at the same subarea with solutions of $\textbf{PS}_t$ \;
   \nl \If{$\textbf{P}_{non}$ is NULL}{
    \textbf{Return}  NULL.
    }
   \nl \Else
  {
  \nl Use the remaining solutions in $\textbf{P}_{non}$ and the extreme point $\textbf{x}_e$ to form a set of lines that represent the directions (denoted as $\textbf{D}$): $\textbf{D}_j = \frac{\textbf{P}^j_{non}-\textbf{x}_e}{\| \textbf{P}^j_{non}-\textbf{x}_e \|}, (j=1,...,|\textbf{P}_{non}|)$\;
  \nl Delete duplicated directions from $\textbf{D}$\;
      }
 \textbf{Return} $\textbf{D}$.
    \caption{{\bf Search of Expansion Direction} \label{Alg:SearchExpanDir}}
\end{algorithm}

Procedures from lines 4 to 10 in Algorithm \ref{Alg:SearchExpanDir} are conducted to make sure the remaining solutions in $\textbf{P}_{non}$ are nondominated and in different sub-spaces from the extreme point such that the extreme point and them can form the right expansion directions. Therefore, in line 4, all solutions in $\textbf{P}_{var}$ are evaluated in the new environment and dominated solutions are discarded after sorting them using nondominated sorting \cite{deb2002fast}. Then, if all solutions in $\textbf{P}_{var}$ are nondominated by those in $PS_t$, just set $\textbf{P}_{non}$ as $\textbf{P}_{var}$ ; else delete all solutions from $\textbf{P}_{var}$ that are dominated by those in $\textbf{PS}_t$ and regard the set of remaining solutions as $\textbf{P}_{non}$ in line 8. Then, in line 9, use evenly generated weight vectors following the method in \cite{chen2017dynamic} to estimate density of $\textbf{P}_{non}$ and $\textbf{PS}_t$ with the method introduced in Section \ref{sec:KTDMOEA_alg}. Delete those solutions from $\textbf{P}_{non}$ that are in the same subarea as those in $\textbf{PS}_t$ in line 10. Later on, if there is no solution in $\textbf{P}_{non}$, this means no expansion direction is found and return NULL; else, in line 13, use the points to form a set of lines that represents the directions (denoted as $\textbf{D}$) by regarding the extreme point $\textbf{x}_e$ in line 2 as the starting point and those solutions in $\textbf{P}_{non}$ as the end point: $\textbf{D}_j = \frac{\textbf{P}^j_{non}-\textbf{x}_e}{\| \textbf{P}^j_{non}-\textbf{x}_e \|}, (j=1,...,|\textbf{P}_{non}|)$. Then, delete duplicated expansion directions from $\textbf{D}$ and return $\textbf{D}$.

After getting the expansion directions, the next step is to expand the PS to generate transferred solutions following the expansion directions. The detailed procedures of this algorithm are shown in Algorithm \ref{Alg:PSExpansion}. Given the Pareto optimal solution set at time t $\textbf{PS}_t$, evenly select solutions from it with the size of $N_{base}$ as $\textbf{P}_{base}$, where
\begin{equation}
N_{base} = \lfloor \frac{N - M_t}{N_{dir} * \theta} \rfloor
\end{equation}
where $N$ is the size of population; $M_{t+1}$ is the NObj at time step t+1; $N_{dir}$ is the number of expansion directions in $\textbf{D}$ and $\theta$ is the number of solutions to generate along each expansion direction, which is a parameter to be set by the user. $N - M_t$ is designed to enable those $M_t$ extreme points in $\textbf{PS}_t$ to be preserved to the next environment. Then in line 3 of Algorithm \ref{Alg:PSExpansion}, generate $\theta$ solutions along each expansion direction in $\textbf{D}$ to fill the transferred solution set $\textbf{P}_{tr}$ through the following Equation (\ref{eq:expand}), which produces a transferred solution based on a base solution and an expansion direction.
\begin{equation}
\begin{split}
\label{eq:expand}
\textbf{x}_{new}^{(j-1)*N_{dir} + i} = \textbf{x}_i + C_i^j * rand() * \textbf{D}_j \\ (i = 1,...,N_{base}; j = 1,...,N_{dir})
\end{split}
\end{equation}
where $\textbf{x}_i$ is the $i$-th solution in the base population $\textbf{P}_{base}$; $C_i^j$ is a float variable enabling some expanded solutions to reach the boundary of the decision space, whose detailed calculation will be discussed in the next paragraph; rand() is a function to generate a random number in $(0,1]$; $\textbf{D}_j$ is the $j$-th expansion direction in the set $\textbf{D}$. After generating transferred solutions through PS expansion, if $\textbf{P}_{tr}$ is not full, just evenly select solutions from $\textbf{PS}_t$ in the objective space with the size of $N - N_{base} * \theta * N_{dir}$.

The calculation of $C_i^j$ should follow the criterion that all solutions expanded from the solutions $\textbf{x}_i$ in $\textbf{P}_{base}$ are within the bound of each decision variable and they should reach the boundary of the search space as close as possible. Bearing this criterion in mind, we design the calculation of $C_i^j$. Given a base solution $\textbf{x}_i$ and one expansion direction $\textbf{D}_j$, suppose $para^k$ is the value that makes the $k$-th variable of the generated solution reach the boundary of this variable. Therefore, each $para^k$ can be calculated according to whether the expansion direction is positive or negative, via the following equation:

\begin{equation}
\label{eq:calculateC}
para^k = \left\{ \begin{array}{ll}
\frac{upper^k - x_i^k}{D_j^k},  & \textrm{$D_j^k > 0$}\\
\frac{lower^k - x_i^k}{D_j^k},  & \textrm{$D_j^k < 0$}\\
\end{array} \right.
\end{equation}
where $upper^k$ and $lower^k$ are the upper bound and lower bound of the $k$-th dimension of the decision space; $x_i^k$ is the $k$-th decision value; $D^k_j$ is the value of the direction $\textbf{D}_j$ at the $k$-th dimension. In order to ensure each generated solution is located within the region, $C_i^j = \underset{k = 1, ..., n}{min}para^k$, where $n$ is the dimension of the decision space.

\subsubsection{Contract the PS when Decreasing the NObj}
\label{ContractPS}

It has been observed that decreasing the NObj usually results in the contraction of the dimension of PS/PF manifold. Therefore, the PS is proposed to be contracted when decreasing the number of objective.

The idea of PS contraction in the decision space is illustrated in Figure \ref{fig:PSContract}. Note this figure is just drawn to demonstrate
the process of PS contraction, the specific cases in real problems may be different. It tries to generate spread  and uniform solutions given current nondominated solution set. Steps from lines 4 to 7 in Algorithm \ref{Alg:PSContraction} are designed to help improving the spread of the population and line 8 tries to make the distribution of population more even. As illustrated in Figure \ref{fig:PSContract}, black start points are the found optimal solutions for the problem with 3 objectives. When decreasing the NObj from 3 to 2, the black start points in the true PS of bi-DMOP (blue line) are selected as the nondominated solutions. 

One solution (denoted by the red point) is generated based on the extreme point (denoted by blue start point) following the spread direction between the extreme point's closest point and itself, so as to increase the spread of population. Other solutions (denoted by the blue points) are produced from two randomly selected solutions in the nondominated solution set, to improve the even distribution of the population (suppose there is no bias in the problem).


\begin{algorithm}[ht!]
\KwIn{Pareto optimal solution set at time t ($\textbf{PS}_t$);}
\KwOut{Transferred solutions $P_{tr}$}
\nl Evaluate the $\textbf{PS}_t$ in the new environment and put the nondominated solutions to $\textbf{P}_{non}$ after conducting the nondominated sorting on $\textbf{PS}_t$\;
  \nl Put all solutions in $\textbf{P}_{non}$ to $\textbf{P}_{tr}$\;
  \nl Find the solutions with the maximum objective value at any objective from $\textbf{P}_{non}$ as the set of extreme points ($\textbf{P}_e$)\;
  \nl \For{j = 1 to $|\textbf{P}_e|$}
  {
    \nl Find a closest solution $\textbf{P}_{non}^j$ to $\textbf{x}_e^j$ from $\textbf{P}_{non}$ and connect $\textbf{P}_{non}^j$ to $\textbf{x}_e^j$ as a direction and normalize it as $\textbf{D}_j = \frac{\textbf{x}_e^j - \textbf{P}_{non}^j}{|\textbf{x}_e^j - \textbf{P}_{non}^j|}$\;
    \nl Produce a new solution $\textbf{x}_{new}$ along the direction $\textbf{D}_j$ to make it reach the boundary of the search space\;
    \nl Put the solution $\textbf{x}_{new}$ to $\textbf{P}_{tr}$\;
  }
  \nl Random select two solutions ($\textbf{x}_a$ and $\textbf{x}_b$) from $\textbf{P}_{tr}$ and generate one solution between them to fill $\textbf{P}_{tr}$ until the size reaches the population size N\;
 \textbf{Return} $\textbf{P}_{tr}$
    \caption{{\bf Contract the PS to generate transferred solutions} \label{Alg:PSContraction}}
\end{algorithm}

Given the Pareto optimal solution set $\textbf{PS}_t$, the first step in line 1 of Algorithm \ref{Alg:PSContraction} is to evaluate all solutions in $\textbf{PS}_t$ in the new environment and put the nondominated solutions to $\textbf{P}_{non}$ after conducting nondominated sorting \cite{deb2002fast} on $\textbf{PS}_t$. Then, put all solutions of $\textbf{P}_{non}$ to the set of transferred solution $\textbf{P}_{tr}$ in line 2. Later on, in line 3, find solutions from $\textbf{P}_{non}$ with the maximum objective value for each objective as the set of extreme points $\textbf{P}_e$. Subsequently, for each extreme point $\textbf{x}_e^j$ in the set $\textbf{P}_e$, find a closest solution $\textbf{P}_{non}^j$ to $\textbf{x}_e^j$ from $\textbf{P}_{non}$ and connect $\textbf{P}_{non}^j$ to $\textbf{x}_e^j$ as a direction and normalize it as $\textbf{D}_j = \frac{\textbf{x}_e^j - \textbf{P}_{non}^j}{|\textbf{x}_e^j - \textbf{P}_{non}^j|}$, as shown in line 5. Then, produce a new solution $\textbf{x}_{new}^i$ along $\textbf{D}_i$ to make it reach the boundary of the search space according to the following equation:
\begin{equation}
\textbf{x}_{new}^j = \textbf{x}_e^j + C_j * \textbf{D}_j
\end{equation}
where $C_j$ is a float variable, whose calculation method is the same as that in the process of PS expansion, as shown in Equation (\ref{eq:calculateC}). The newly generated solution $\textbf{x}_{new}$ is then put in $\textbf{P}_{tr}$. Later on, in line 8, randomly select two solutions ($\textbf{x}_a$ and $x_b$) from $P_{tr}$ and generate one solution between them through equation (\ref{eq:uniform}) to fill $P_{tr}$ until the size of $P_{tr}$ reaches the population size $N$, using the following equation:
\begin{equation}
\label{eq:uniform}
\textbf{x}_{new} = \textbf{x}_a + rand() * (\textbf{x}_a - \textbf{x}_b)
\end{equation}

We believe this strategy  of expanding/contracting the PF and PS works better over DTAEA \cite{chen2017dynamic} because increasing or decreasing the NObj usually results in the expansion and contraction of the dimension of the PS manifold. When increasing the NObj, the proposed PS expansion is able to find the expansion directions and generate solutions along these directions, increasing the population diversity in the new environment. When decreasing the NObj, the two mechanisms in PS contraction are targeted for improving the spread and evenness of the distribution. Therefore, those produced solutions by PS expansion and contraction may achieve better diversity than that of DTAEA.

\subsection{KTDMOEA}
\label{sec:KTDMOEA_alg}

As the PS expansion/contraction is designed to enhance the diversity of population after changes, the proposed KTDMOEA will not maintain a separate DA. As a result, there is no need to update DA as in DTAEA. The population update mechanism in KTDMOEA is the same as the update mechanism of CA in \cite{chen2017dynamic}.

\begin{algorithm}[ht!]
\KwIn{Population size $N$;}
\KwOut{The found population $Pop$}
\nl Randomly generate an initial population $Pop$\;
\nl \While{stopping criteria not satisfied}
  {
  	\nl \If{the number of objective changes}
  	{
  \nl \If{the number of objective increases}{
  \nl Conduct the process of PS expansion in Algorithm \ref{Alg:PSExpansion} \;}
  \nl \ElseIf{the number of objective decreases}
  {
  \nl Conduct the process of PS contraction in Algorithm \ref{Alg:PSContraction}\;
      }
  }
  \nl \Else
  {
  \nl Generate an offspring population $A$ using the mating selection and genetic operators based on the parent population $Pop$\;
  \nl Update $Pop$ using offspring population $A$ with the CA update mechanism in DTAEA \cite{chen2017dynamic}  \;
      }
  }
 \textbf{Return} $Pop$
    \caption{{\bf Framework of KTDMOEA} \label{Alg:KTDMOEA}}
\end{algorithm}

The flowchart of KTDMOEA is shown in Figure \ref{fig:flowKTMOEA}. The overall framework of the proposed KTDMOEA is given in Algorithm \ref{Alg:KTDMOEA}. KTDMOEA starts with generating an initial population of size $N$, as shown in line 1. While the stopping criteria are not satisfied, carry out the following steps. Detect whether the environmental changes occur. If the NObj is detected to increase, evoke the process of PS expansion on $Pop$ using Algorithm \ref{Alg:PSExpansion}. If the NObj decreases, evoke the process of PS contraction on $Pop$ using Algorithm \ref{Alg:PSContraction}. If there is no change detected, conduct the evolutionary optimization process on $Pop$. In line 9, an offspring population $A$ is produced through the following two steps until the size of $A$ reaches the population size $N$. Firstly, randomly pick two solutions from $Pop$ as the parent solutions via the mating selection. Then, those two solutions are used to generate two offspring solutions via appropriate crossover and mutation operators. Here, we utilize the simulated binary crossover \cite{deb1995simulated} and the polynomial mutation, as used by most continuous multi-objective evolutionary algorithms including DTAEA.

Lastly, the generated offspring population $A$ is used to update $Pop$ with the CA update mechanism in DTAEA \cite{chen2017dynamic}. Due to space limitation, the CA update mechanism of DTAEA is not introduced here. Interested readers can refer to \cite{chen2017dynamic}.

\vspace{0.6cm}

\section{Experimental Setup}
\label{sec:ExperSetup}

In this section, experimental studies are designed to verify whether the improved knowledge transfer-based approach answers the research questions mentioned in Section \ref{sec:Intro}. Analyses will be carried out to reveal whether existing DMOEAs for DMOPs are able to deal with a changing NObj despite not being designed to do so, and how well static MOEAs could perform on DMOPs with a changing NObj. These are important baselines.

\subsection{Benchmark Problems}
\label{sec:prob}

Two suites of multi-objective optimization test problems  DTLZ \cite{deb2005scalable} and WFG \cite{huband2006review} are modified to be DMOPs with a changing NObj. Four DMOPs with a changing NObj from DTLZ1-DTLZ4 are renamed as F1-F4, the same as in \cite{chen2017dynamic}. These two suites of benchmark functions are used to verify that the proposed algorithm is able to deal with problems with both simple and complex problem features. Detailed descriptions of problems features can be found in Section I of our Supplementary File, which can be download from {https://github.com/ganrandom/Supplementary-File-for-KTDMOEA}.

There are two different sequences of changes for these benchmark problems:
\begin{enumerate}
\item The initial NObj is set as 2. It firstly increases from 2 to 7 one by one and then decreases from 7 to 2 one by one (simply denoted as `2-7-2 one by one'), which was used in \cite{chen2017dynamic};
%
\item The initial NObj is set as 7. It firstly decreases from 7 to 2 one by one and then increases from 2 to 7 one by one (simply denoted as `7-2-7 one by one').
%

\end{enumerate}

\subsection{Compared Algorithms}

In our experimental studies, five algorithms are selected for the comparison, so as to verify the performance of our proposal against the state-of-the-art. Considering the popularity and good performance on solving static MOPs, the elitist nondominated sorting genetic algorithm (NSGA-II) \cite{deb2002fast} and multi-objective evolutionary algorithm based on decomposition (MOEA/D) \cite{zhang2007moea} are chosen to verify whether they are able to tackle DMOPs with a changing NObj. For NSGA-II and MOEA/D, whenever there is a change, the whole population of the last generation in the old environment is just copied to the next generation after changes and then re-evaluated in the new environment to respond to changes in the NObj. Besides, two popular and state-of-the-art DMOEAs including DNSGA-II \cite{application_scheduling_deb2007dynamic} and MOEA/D-KF \cite{muruganantham2015evolutionary} specifically designed for DMOPs with a changing shapes and/or positions of PS and/or PF are selected to be compared, so as to figure out whether DMOEAs for solving DMOPs with fixed NObj are able to deal with DMOPs with a changing NObj. These four algorithms are compared to verify whether it is necessary to develop extra approaches tailored for DMOPs with a changing NObj. In order to verify whether the improved knowledge transfer-based approach (KTDMOEA) answers the research questions mentioned in Section \ref{sec:Intro}, as one of the popular and recently developed algorithms targeted for handling changes in the NObj, DTAEA \cite{chen2017dynamic} is also chosen to be compared. Section II of the supplementary file presents the detailed descriptions of these algorithms.

\subsection{Parameter Settings}
\label{sec.Para}
The parameters of these compared algorithms are set as follows:
\begin{itemize}
\item Population size: 300, the same as that of DTAEA, $\theta$ in KTDMOEA is set as 2. The impact of $\theta$ on KTDMOEA's performance will be analyzed in Section \ref{sec.ImpactAlgPara};
\item Several different frequencies of change: $\tau_ t$ is set as 5, 25 and 50 and 200; Those parameters are set for assessing the effects of different algorithms under different frequencies of change.
\item All algorithms run 31 times independently, also the same as in DTAEA's work \cite{chen2017dynamic};
\item 1000 generations are given to each algorithm before the first change so that the population before the change can converge;
\item The crossover probability was $p_c$ = 1.0
and its distribution index was $\eta _c$ = 20. The mutation probability was $p_m$ = 1/n (where n denotes the number of decision variables) and its distribution $\eta _m$ = 20. These parameters are chosen because of their good performance on solving continuous problems, which have been analyzed in \cite{2001_Deb_book_multiobjectives} and \cite{deb2005scalable}.
\item The neighbourhood size and the number $n_r$ of solutions allowed to replace in MOEA/D were set to 20 and 2, respectively, which is the same as in the original paper \cite{zhang2007moea}.
\end{itemize}

\subsection{Performance Metrics}
\begin{itemize}
\item Hypervolume (HV) \cite{zitzler2003performance} comprehensively measures the convergence and diversity of solution sets; the larger the better.
\item Generational Distance (GD) \cite{goh2009competitive} \cite{ruan2017effect} evaluates the convergence of obtained solution sets; the smaller the better.
\item Maximum Spread (MS) \cite{li2019quality} assesses the diversity of solution sets; the larger the better.
\end{itemize}
Note that these three metrics are used to measure the solution quality of a found solution set. They can be also used to measure comprehensive performance of an algorithm by averaging the metric values of all obtained solutions under multiple environmental changes.

%
%
%
%
%
%


\section{Experimental Results and Analyses}
\label{sec:results}

In order to achieve the objectives of the experiment, i.e. answering the research questions and verifying whether the existing static MOEAs and DMOEAs for solving DMOPs with fixed NObj are able to tackle DMOPs with a changing NObj, experimental results of all compared algorithms are presented in this section. Furthermore, further analyses regarding the further verification of the improved knowledge transfer, performance comparison of different NObj changing sequence and the impact of algorithm parameters are also given in this section.

Three metrics (HV, GD and MS) are used to measure the quality of the found solutions at the first generation after the change and in the last generation before the next change by six algorithms. To show the significant superiority of the proposed KTDMOEA to other algorithms across all problem instances in general, Friedman and Nemenyi statistical tests \cite{demvsar2006statistical} are adopted across all benchmark problems regarding the three metrics (HV, GD and MS) of six compared algorithms. The larger the values of HV and MS, the better the algorithm. Therefore, the larger the Friedman ranking, the better the algorithm. Similarly, the smaller the Friedman ranking of GD, the better the algorithm.

The mean metric value of 31 independent runs that each algorithm gets on one problem with one frequency of change at each environmental change is regarded as an observation of the Friedman and Nemenyi test. Therefore, there are 520 (13 problems, 4 different frequencies of change and 10 environmental changes) observations for each algorithm in the Friedman and Nemenyi tests. Additionally, in order to show the significant superiority of the proposed KTDMOEA to other algorithms on each individual problem of each parameter setting, the Wilcoxon rank sum test at the 5\% significance level is implemented on each  benchmark problem regarding each metric of six compared algorithms at each parameter setting. Therefore, there are 31 observations obtained from 31 independent runs for each algorithm on each problem and parameter setting in the Wilcoxon rank sum test.

Due to the space limitation of the paper, only the results of the Friedman and Nemenyi statistical tests are presented here. The Wilcoxon rank sum test results are shown in the Supplementary file. Mean and standard deviation values of HV, GD and MS of obtained solutions in the first generation after changes and the last generation before the next change averaged across 10 environmental changes in two sequences of changes as `2-7-2 one by one' 
and `2-7-2 one by one' are also presented in the supplementary file, respectively. Moreover, mean and standard deviation values of HV, GD and MS of obtained solutions at the first generation after changes and at the last generation before the next change at each environmental changes under 31 independent runs in those two sequences of changes as `2-7-2 one by one' and `2-7-2 one by one' are also recorded and presented in the supplementary file.

\subsection{Initial Effectiveness of Knowledge Transfer}
\label{sec:InitialKT}

In order to verify (1) whether the proposed PS expansion/contraction mechanism is able to increase diversity so as to improve knowledge transfer after the change, and (2) whether the MOEAs not tailored for DMOPs with a changing NObj can achieve this aim after the change, the quality of the solutions obtained by all algorithms in the first generation after changes is compared.

\subsubsection{NObj increasing from 2 to 7 and then decreasing from 7 to 2}
\label{sec:1st272}

\begin{figure}
\hspace{-.3in}
  \centering
  \subfigure[HV]{
    \label{fig:HV} 
    \includegraphics[width=0.35\linewidth]{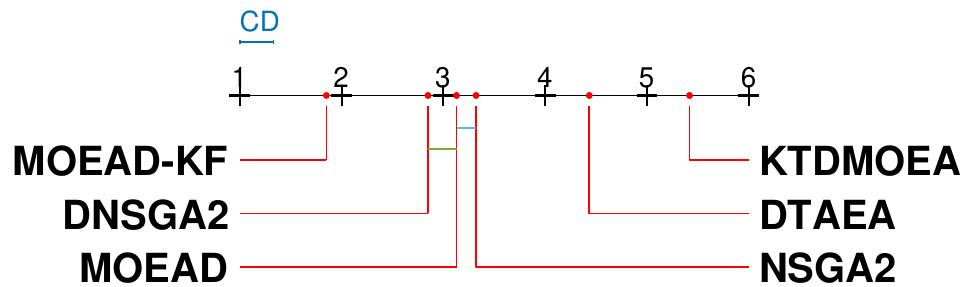}}
  \hspace{-.14in}
  \subfigure[GD]{
    \label{fig:GD} 
    \includegraphics[width=0.35\linewidth]{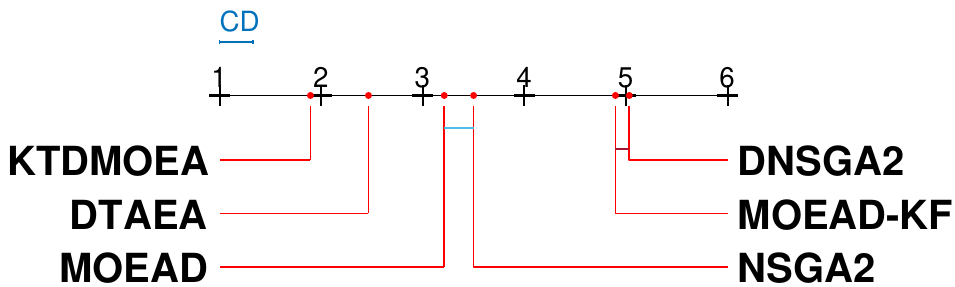}}
    \hspace{-.14in}
	\subfigure[MS]{
    \label{fig:MS} 
    \includegraphics[width=0.35\linewidth]{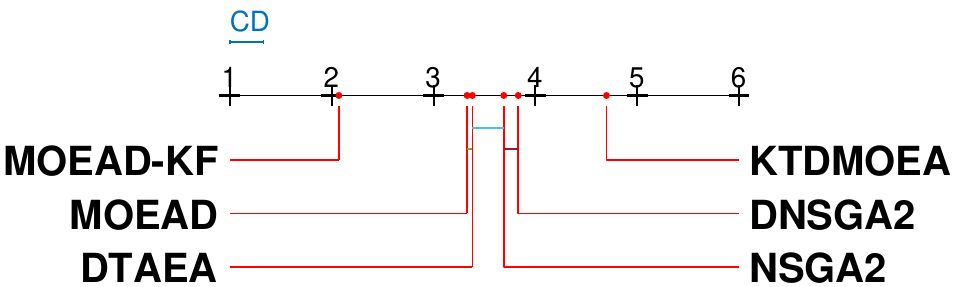}}
    \hspace{-.3in}
    \vspace{-0.3cm}
  \caption{Friedman ranking among HV, GD and MS of obtained solutions at the first generation by 6 algorithms in the changing sequence of \textbf{firstly increasing the NObj from 2 to 7 and then decreasing it from 7 to 2}, both one by one.}
  \label{fig:1st272} 
  \vspace{-0.4cm}
\end{figure}

Figure \ref{fig:1st272} presents the Nemenyi post-tests results among HV, GD and MS of obtained solutions at the first generation after changes by 6 algorithms. Friedman test detects significant differences in average values for HV, GD and MS with a p-value of 3.57E-251, 9.14E-256, and 1.44E-117, respectively.

Overall, it can be observed from Figure \ref{fig:1st272} that when comparing all algorithms, KTDMOEA significantly outperforms all others in all three metrics. More details can be found from Table 2 of the Suppmentary File. This implies that the proposed knowledge transfer technique via PS expansion/contraction indeed improves the diversity and maintain the convergence of transferred solutions right after changes, under all frequencies of changes on most problems.

For readers who want to examine the details, results of mean and standard deviation values for HV, GD and MS when the NObj increase from 2 to 7 and then decrease from 7 to 2 are presented in Tables 6, 7 and 8 of the Supplementary File, respectively. 

\subsubsection{NObj decreasing from 7 to 2 and then increasing from 2 to 7}

\begin{figure}
\hspace{-.3in}
  \centering
  \subfigure[HV]{
    \label{fig:HV} 
    \includegraphics[width=0.35\linewidth]{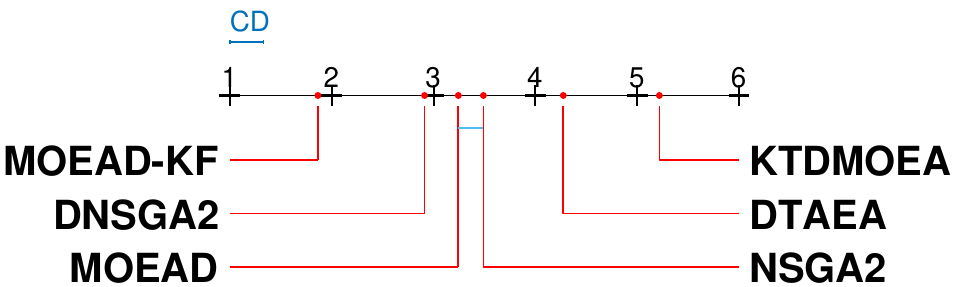}}
  \hspace{-.14in}
  \subfigure[GD]{
    \label{fig:GD} 
    \includegraphics[width=0.35\linewidth]{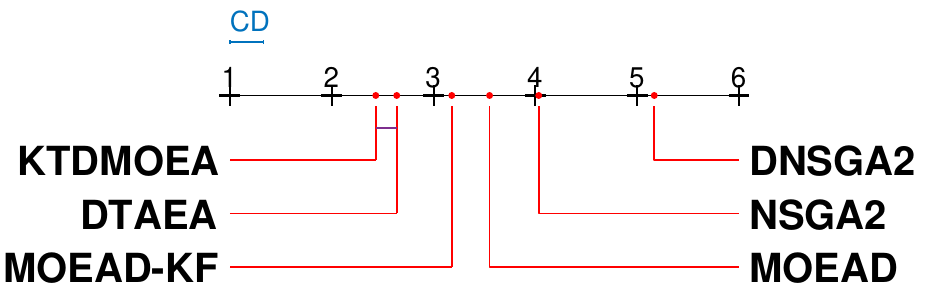}}
    \hspace{-.14in}
	\subfigure[MS]{
    \label{fig:MS} 
    \includegraphics[width=0.35\linewidth]{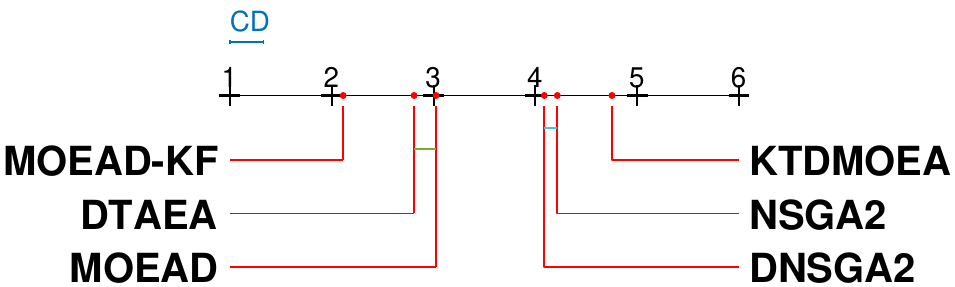}}
    \hspace{-.3in}
    \vspace{-0.3cm}
  \caption{Friedman ranking among HV, GD and MS of obtained solutions at the first generation by 6 algorithms in the changing sequence of \textbf{firstly decreasing the NObj from 7 to 2 and then increasing it from 2 to 7}, both one by one.}
  \label{fig:1st727} 
  \vspace{-0.4cm}
\end{figure}

Figure \ref{fig:1st727} presents the Nemenyi post-tests results among HV, GD and MS of obtained solutions at the first generation by 6 algorithms. Friedman test detects significant differences in average accuracy for HV, GD and MS with a p-value of 1.97E-240, 4.72E-221, and 6.87E-113, respectively.

Overall, it can be found from the Friedman test results in Figure \ref{fig:1st727} that KTDMOEA performs significantly better than all others regarding HV and MS metrics. There is no significant difference between KTDMOEA and DTAEA regarding GD. More details can be found from Table 3 of the Suppmentary File. This further supports  that the proposed knowledge transfer technique via PS expansion/contraction indeed improves the diversity and maintains the convergence of transferred solutions right after changes, under all frequencies of changes on most problems.

For readers who are interested in details, mean and standard deviation values for HV, GD and MS in the benchmark of decreasing the NObj from 7 to 2 and then increasing it from 2 to 7 are presented in Tables 9, 10 and 11 of the Supplementary File, respectively. The comparison results of all algorithms at each NObj regarding HV, GD and MS are presented in Tables 44-53, Tables 54-63 and Tables 64-73 of the Supplementary File, respectively.

\subsubsection{Why Does Knowledge Transfer Usually Get Better Solution Quality Right after Changes?}

In this section, two examples are presented to elaborate the reason why the proposed PS expansion/contraction works well on most problems.

\begin{figure}
  \centering
  \subfigure[F2]{
    \label{fig:ExpandDTLZ2} 
    \includegraphics[width=0.48\linewidth]{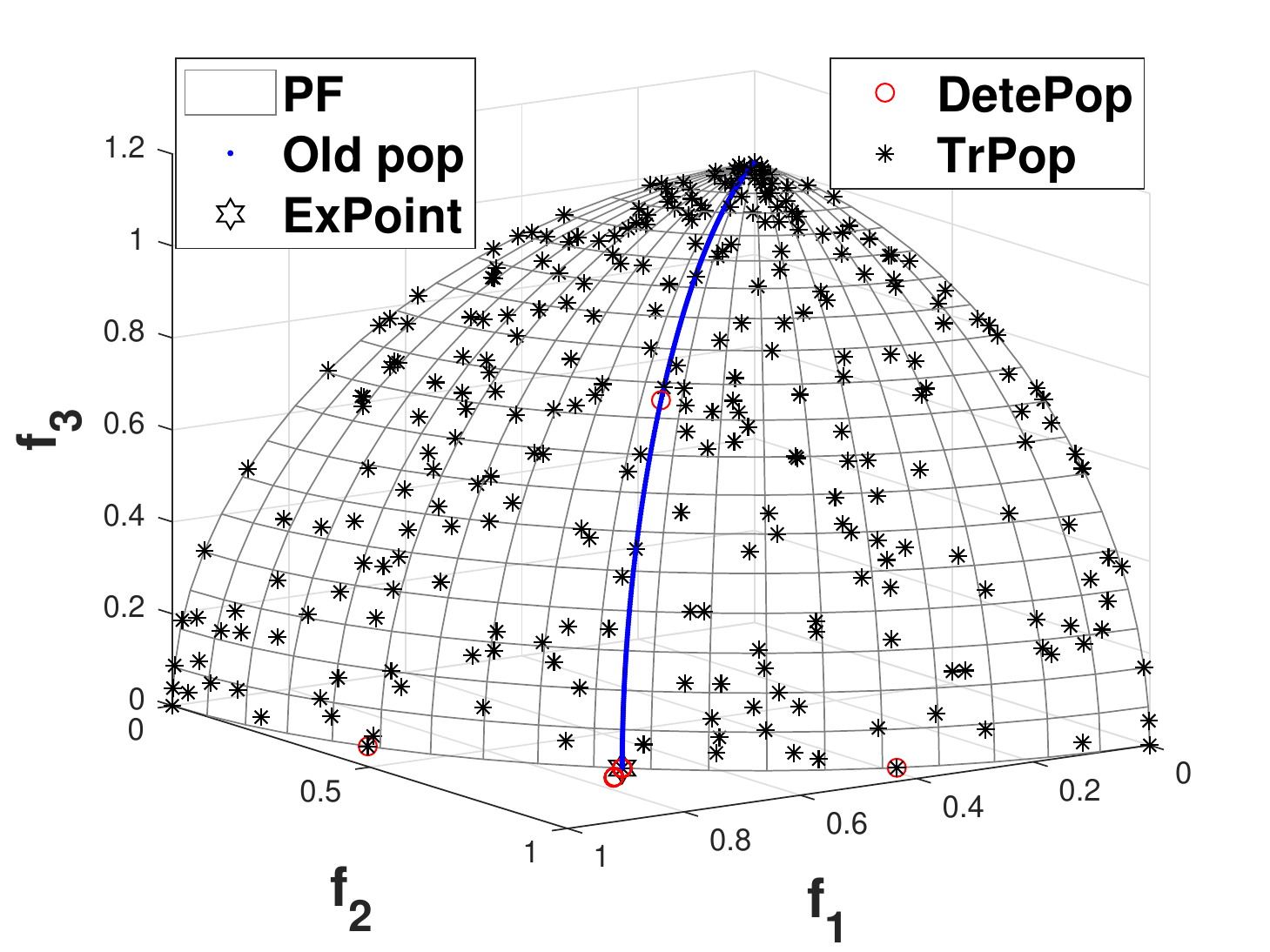}}
  \subfigure[WFG1]{
    \label{fig:ExpandWFG1} 
    \includegraphics[width=0.48\linewidth]{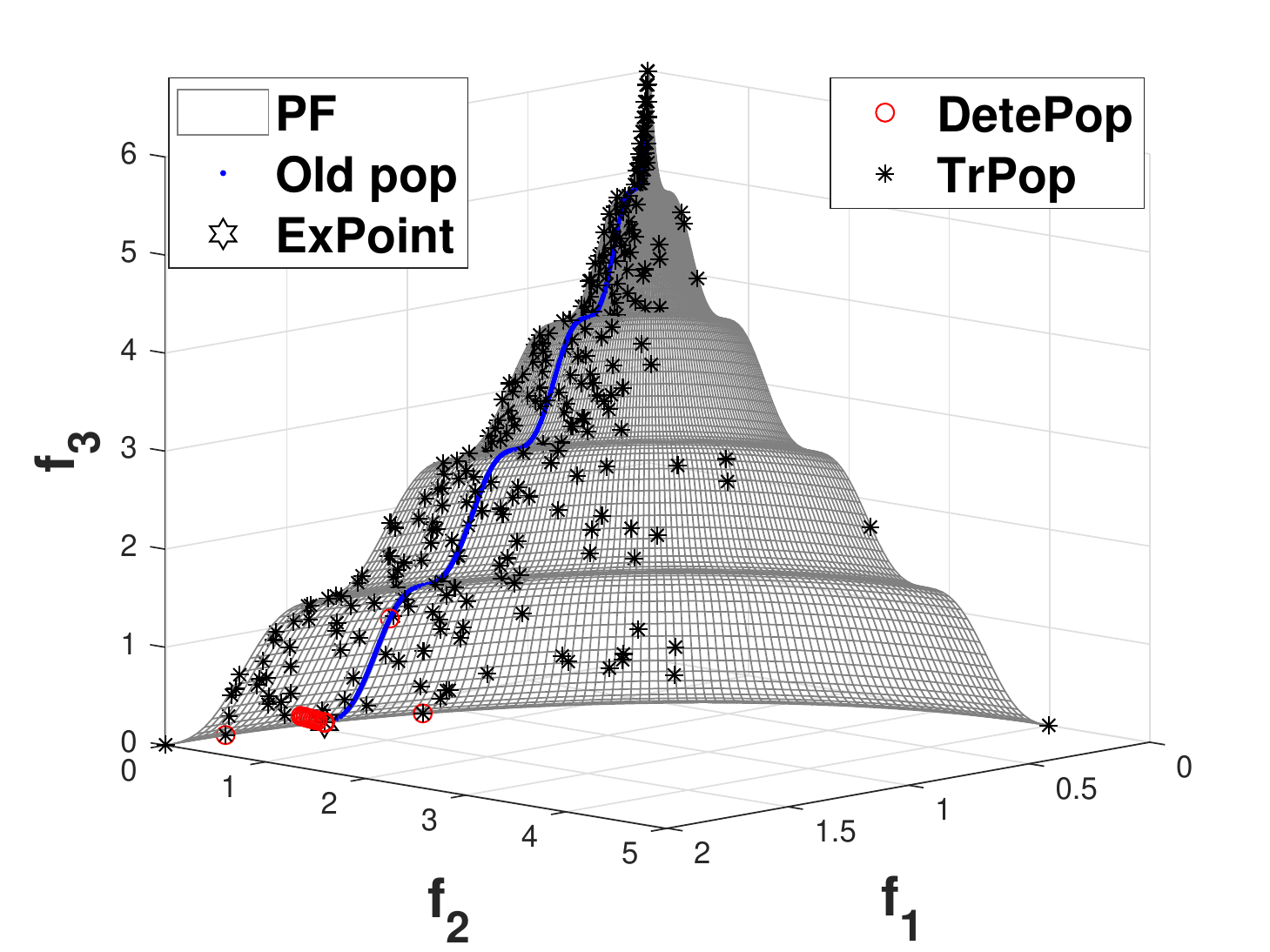}}\vspace{-0.35cm}
  \caption{The distribution of the old population (`Old pop'), the extreme point (`ExPoint'), the detective population (`DetePop') and transferred population (`TrPop') via PS expansion on F2 and WFG1 at the first generation when increasing the NObj from 2 to 3.}
  \label{fig:expandPlot} 
  \vspace{-0.4cm}
\end{figure}

As shown in Figure \ref{fig:expandPlot}, the distributions of the old population, the detective population and transferred population via PS expansion on F2 and WFG1 is presented when increasing the NObj from 2 to 3. It is clear that nondominated solutions in the detective population are still nondominated by the selected extreme point. Following steps 5 to 10 in Algorithm \ref{Alg:SearchExpanDir}, there are only two solutions in $P_{var}$, which are located in the areas away from that of the old population. Then, each of those two solutions is regarded as the ending point of the expansion direction, together with the extreme point as the starting point of the direction. Therefore, when evenly selecting solutions from the old population to conduct the PS expansion, almost all areas of F2 can be covered right after the change. Even for WFG1, a large area of the PF is covered by the transferred solutions. In addition, it is clear that some of the transferred solutions are able to reach the boundary of the PF.

\begin{figure}
  \centering
  \subfigure[F2]{
    \label{fig:ContractDTLZ2} 
    \includegraphics[width=0.48\linewidth]{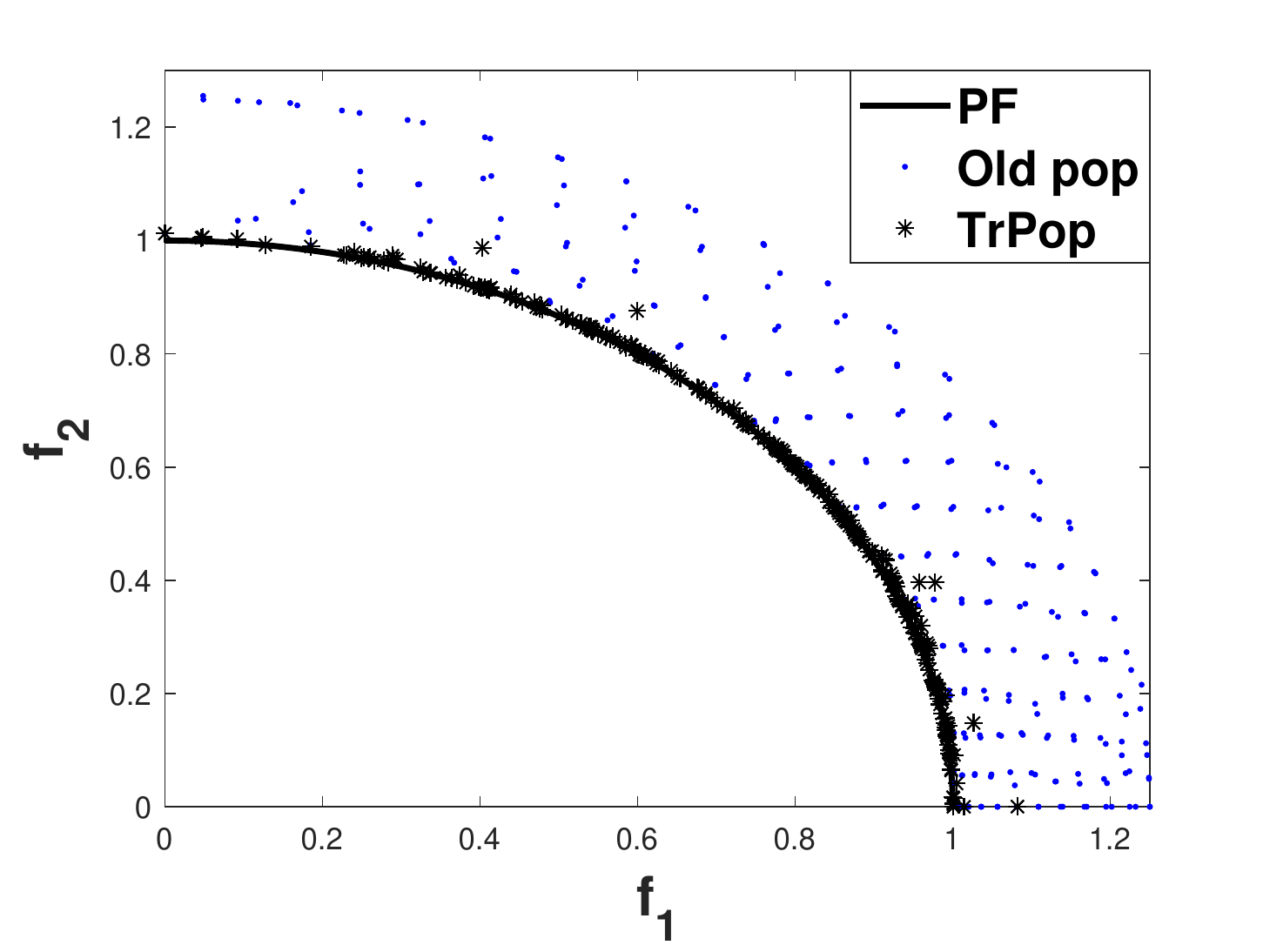}}
  \subfigure[WFG6]{
    \label{fig:ContractWFG6} 
    \includegraphics[width=0.48\linewidth]{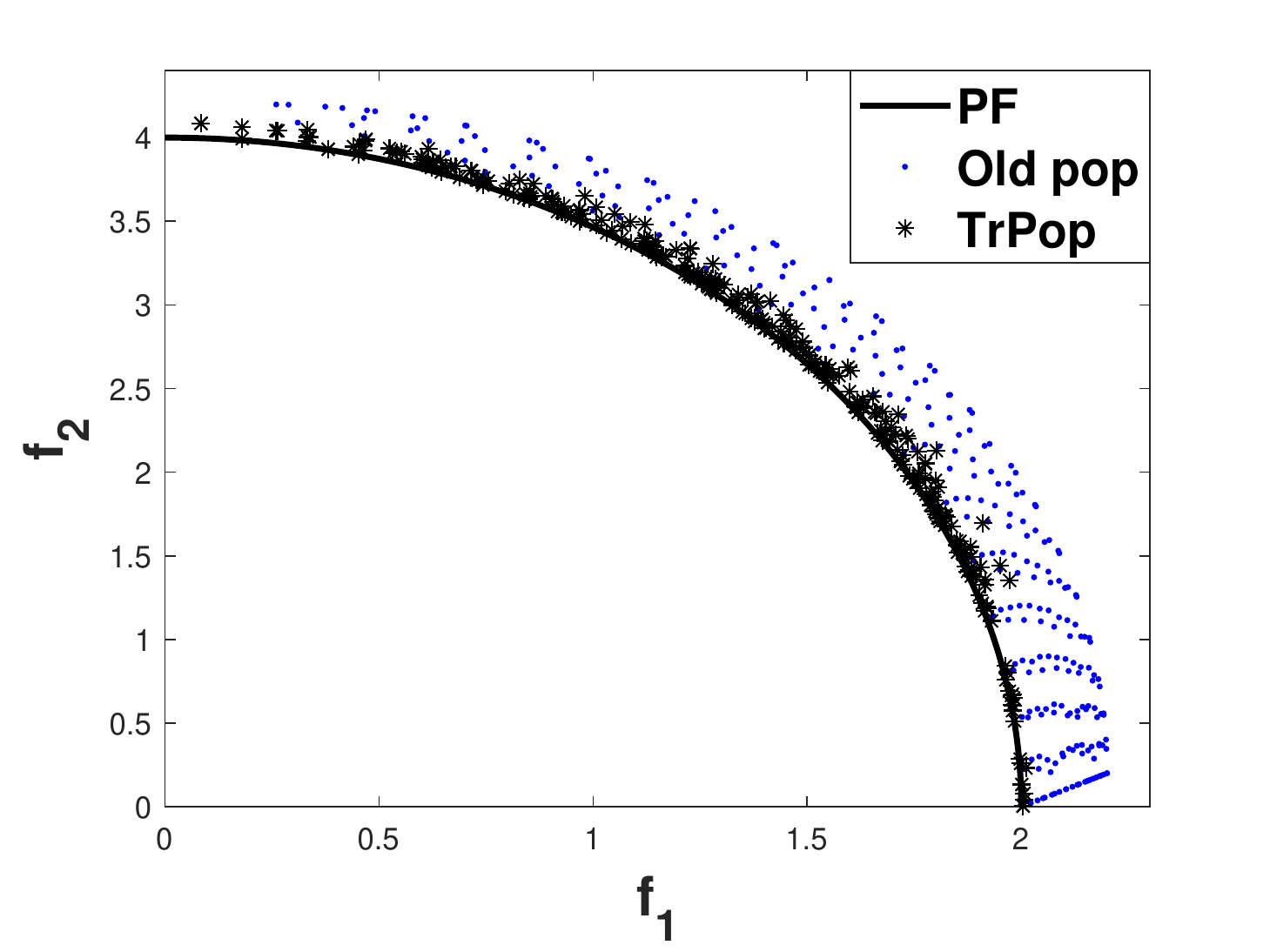}}\vspace{-0.35cm}
  \caption{The distribution of the old population (`Old pop') and transferred population (`TrPop') via PS contraction on F2 and WFG1 at the first generation when decreasing the NObj from 3 to 2.}
  \label{fig:contractPlot} 
  \vspace{-0.6cm}
\end{figure}

Figure \ref{fig:contractPlot} presents the distributions of the old population and the transferred solutions via PS contraction on F2 and WFG1 at the first generation when decreasing the NObj from 3 to 2 in the changing NObj sequence of firstly decreasing from 7 to 2 and then increasing from 2 to 7. It is clear that on those two problems the transferred population via PS contraction has better convergence and diversity than the old population.

\subsection{How Does Knowledge Transfer Help Optimization?}

In order to verify whether the proposed KTDMOEA can find  solutions with better convergence and diversity in the last generation after optimization against all other algorithms, the solution quality of all compared algorithms after optimization and before the next change is compared.

\subsubsection{NObj increasing from 2 to 7 and then decreasing from 7 to 2}
\label{sec:last272}

\begin{figure}
\vspace{-0.5cm}
\hspace{-.3in}
  \centering
  \subfigure[HV]{
    \label{fig:lastHV272} 
    \includegraphics[width=0.35\linewidth]{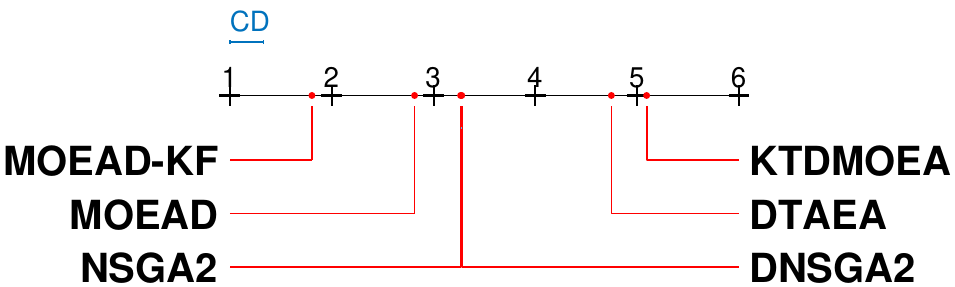}}
  \hspace{-.14in}
  \subfigure[GD]{
    \label{fig:lastGD272} 
    \includegraphics[width=0.35\linewidth]{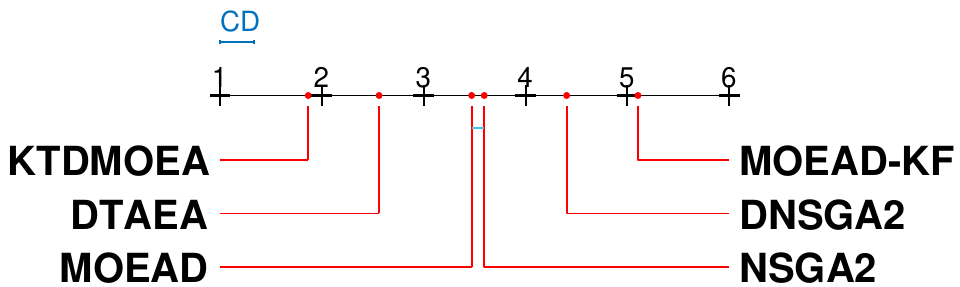}}
    \hspace{-.14in}
	\subfigure[MS]{
    \label{fig:lastMS272} 
    \includegraphics[width=0.35\linewidth]{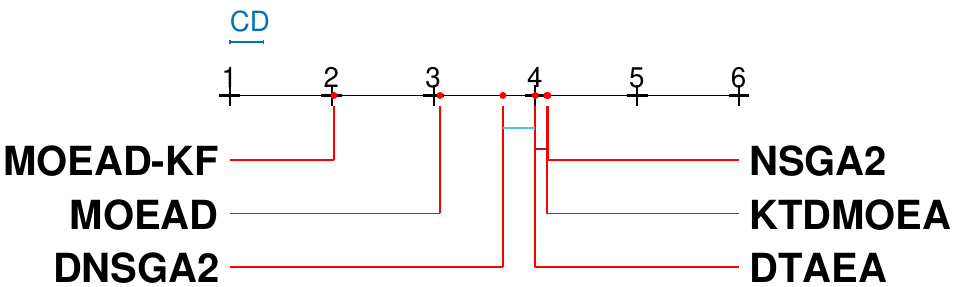}}
    \hspace{-.3in}
    \vspace{-0.3cm}
  \caption{Friedman ranking among HV, GD and MS of optimized solutions at the last generation by 6 algorithms in the changing sequence of \textbf{firstly increasing the NObj from 2 to 7 and then decreasing it from 7 to 2}, both one by one.}
  \label{fig:last272} 
  \vspace{-0.6cm}
\end{figure}

Figure \ref{fig:last272} presents the Nemenyi post-tests results among HV, GD and MS of obtained solutions at the last generation after optimization by 6 algorithms.
Friedman test detects significant differences in average accuracy for HV, GD and MS with a p-value of 6.83E-215, 6.22E-160, and 4.27E-164, respectively.

Overall, it can be seen from those statistical test results that KTDMOEA performs significantly better than or the same as the other approaches. Specifically, it is clear from the Friedman ranking results in Figure \ref{fig:last272} that KTDMOEA gets significantly best results among all compared algorithms regarding HV and GD values. It is the equal best, together with DTAEA and NSGA2, regarding the MS value. These three algorithms outperforms other algorithms regarding the MS value. The statistical results show that the proposed knowledge transfer is able to help the optimization, which achieves better convergence and at least similar diversity compared to the start-of-the-arts when the NObj increasing from 2 to 7 and then decreasing from 7 to 2, under all frequencies of changes on most problems. More details can be seen from Table 4 in the Supplementary File. 

\subsubsection{NObj decreasing from 7 to 2 and then increasing from 2 to 7}
\label{sec:last727}

\begin{figure}
\hspace{-.3in}
  \centering
  \subfigure[HV]{
    \label{fig:lastHV727} 
    \includegraphics[width=0.35\linewidth]{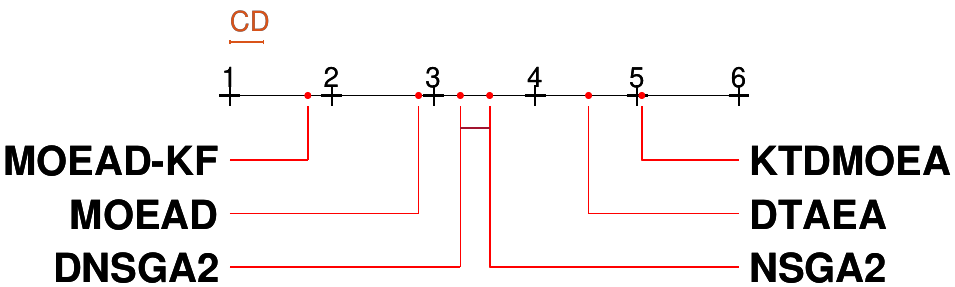}}
  \hspace{-.14in}
  \subfigure[GD]{
    \label{fig:lastGD727} 
    \includegraphics[width=0.35\linewidth]{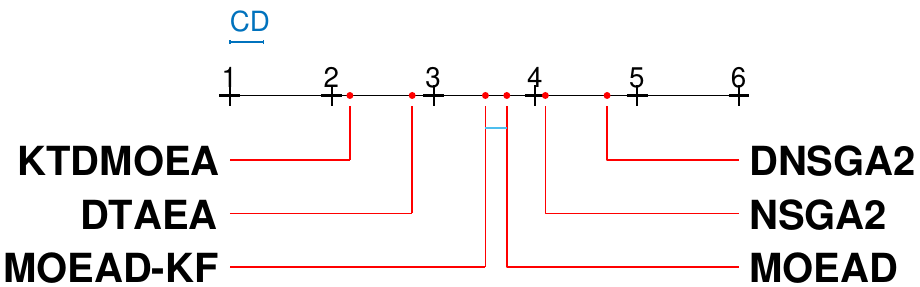}}
    \hspace{-.14in}
	\subfigure[MS]{
    \label{fig:lastMS727} 
    \includegraphics[width=0.35\linewidth]{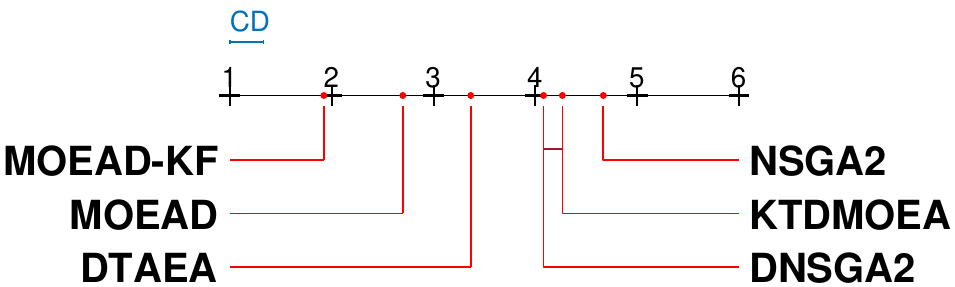}}
    \hspace{-.3in}
    \vspace{-0.3cm}
  \caption{Friedman ranking among HV, GD and MS of optimized solutions at the last generation by 6 algorithms in the changing sequence of \textbf{firstly decreasing the NObj from 7 to 2 and then increasing it from 2 to 7}, both one by one.}
  \label{fig:last727} 
  \vspace{-0.4cm}
\end{figure}

Figure \ref{fig:last727} presents the Nemenyi post-tests results among HV, GD and MS of obtained solutions at the last generation after optimization by 6 algorithms. Friedman test detects significant differences in average values for HV, GD and MS with a p-value of 3.56E-223, 1.58E-129, and 4.98E-183, respectively.

It can be found from the Friedman test results in Figure \ref{fig:last727} that KTDMOEA achieves significantly better results than all other algorithms regarding HV and GD metrics. As for the MS results, KTDMOEA and DNGSA2 rank the second in the Friedman ranking test, both are outperformed by NSGA2 only. Overall, those statistical results imply that the proposed knowledge transfer is able to help the optimization in obtaining better convergence and at least similar diversity compared to the start-of-the-arts in the changing sequence of firstly decreasing from 7 to 2 and then increasing from 2 to 7, under all frequencies of changes on most problems.

\subsubsection{Why Does Knowledge Transfer Help Optimization?}
It has been presented in Section \ref{sec:InitialKT} that the proposed PS expansion/contraction has indeed enhanced the diversity of knowledge transfer, resulting in better solution quality of obtained solutions than other state-of-the-arts in the first generation after changes. 
In other words, given the results of better solution quality than other algorithms in the first generation after changes, our proposed KTDMOEA is able to find solutions with good
convergence and diversity at at all frequencies of change, even when the frequency of change is very
high. This means that our proposed approach is robust to different frequencies of change.

Because the transferred solutions are better distributed in the new environment with a better diversity already, KTDMOEA is able to find better solutions across different frequencies of change. This is also the reason why KTDMOEA is able to quickly respond to the changes in the NObj, since finding good
solution under high frequency of change means fast response to changes. There are some problems where KTDMOEA did not perform best when the frequency of changes is large. The specific results and analyses are presented in Section III.B.3) of the Supplementary File.

\subsection{Further Analysis of Our Knowledge Transfer Methods}
In order to further verify the effectiveness of the proposed PS expansion/contraction method against the state-of-the-art method DTAEA, two pairs of comparisons are designed. First, DTAEA is compared to DTAEAv1, where the CA reconstruction after changes is replaced by the proposed PS expansion/contraction with other components of DTAEA unchanged. Second, KTDMOEA is compared to KTDMOEAv1, where the proposed PS expansion/contraction is replaced by the CA reconstruction of DTAEA with other components of KTDMOEA unchanged.

All experimental settings are set the same as in Section \ref{sec.Para} except for the frequency of change and the NObj changing sequence, which is set as 25 and NObj increasing from 2 to 7 and then decreasing from 7 to 2 one by one, respectively, to save space. For Friedman and Nemenyi tests, the mean metric value of 10 environmental changes that each algorithm gets on one problem with one frequency of change at each independent run is regarded as an observation of the test. Therefore, there are 403 (13 problems and 31 environmental changes) observations for each algorithm in the Friedman and Nemenyi tests.

\begin{figure}
\hspace{-.3in}
  \centering
  \subfigure[HV]{
    \label{fig:DTAEAHV} 
    \includegraphics[width=0.35\linewidth]{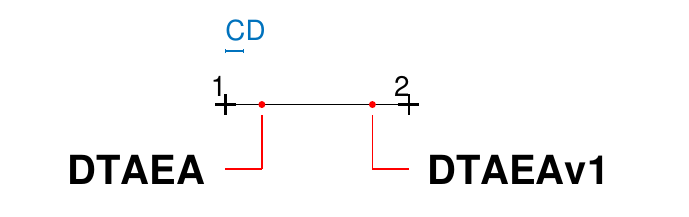}}
  \hspace{-.14in}
  \subfigure[GD]{
    \label{fig:DTAEAGD} 
    \includegraphics[width=0.35\linewidth]{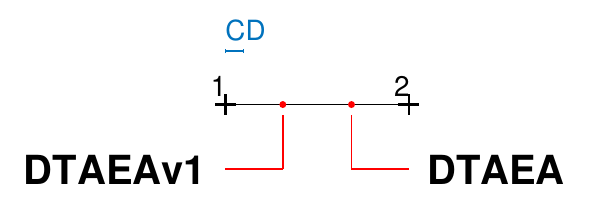}}
    \hspace{-.14in}
	\subfigure[MS]{
    \label{fig:DTAEAMS} 
    \includegraphics[width=0.35\linewidth]{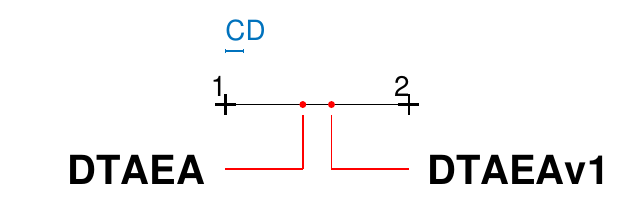}}
    \hspace{-.3in}
    \vspace{-0.3cm}
  \caption{Friedman ranking among HV, GD and MS of optimized solutions at the last generation by DTAEA and DTAEAv1 in the changing sequence of \textbf{firstly increasing the NObj from 2 to 7 and then decreasing it from 7 to 2}, both one by one.}
  \label{fig:FurtherAnaDTAEA} 
  \vspace{-0.4cm}
\end{figure}
\begin{figure}
\hspace{-.3in}
  \centering
  \subfigure[HV]{
    \label{fig:KTHV} 
    \includegraphics[width=0.35\linewidth]{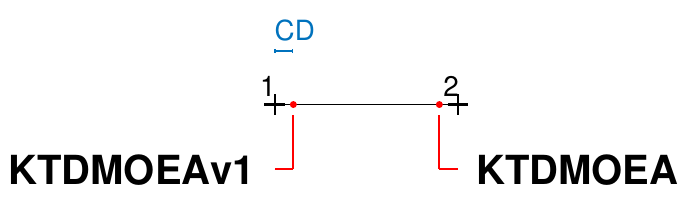}}
  \hspace{-.14in}
  \subfigure[GD]{
    \label{fig:KTGD} 
    \includegraphics[width=0.35\linewidth]{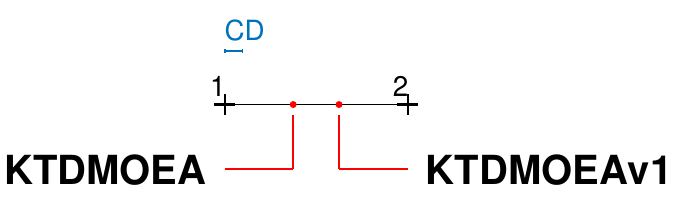}}
    \hspace{-.14in}
	\subfigure[MS]{
    \label{fig:KTMS} 
    \includegraphics[width=0.35\linewidth]{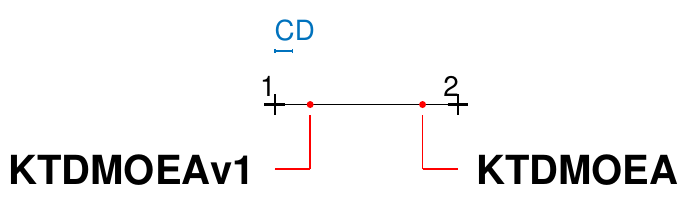}}
    \hspace{-.3in}
    \vspace{-0.3cm}
  \caption{Friedman ranking among HV, GD and MS of optimized solutions at the last generation by KTDMOEA and KTDMOEAv1 in the changing sequence of \textbf{firstly increasing the NObj from 2 to 7 and then decreasing it from 7 to 2}, both one by one.}
  \label{fig:FurtherAnaKT} 
  \vspace{-0.4cm}
\end{figure}

Figure \ref{fig:FurtherAnaDTAEA} presents the Nemenyi post-tests results among HV, GD and MS of obtained solutions at the last generation after optimization before the next change by DTAEA and DTAEAv1 algorithms. Friedman test detects significant differences in average values for HV, GD and MS with a p-value of 9.97E-34, 5.40E-14, 0.0017, respectively. Similarly, Figure  \ref{fig:FurtherAnaKT} presents the Nemenyi post-tests results among HV, GD and MS of obtained solutions at the last generation after optimization before the next change by KTDMOEA and KTDMOEAv1 algorithms. Friedman test detects significant differences in average values for HV, GD and MS with a p-value of 1.50E-57, 4.88E-07, 8.63E-35, respectively.

Overall, it can be observed from Figures \ref{fig:FurtherAnaDTAEA} and \ref{fig:FurtherAnaKT} the algorithm with our proposed knowledge transfer strategy significantly outperforms the one without it, i.e, DTAEAv1 outperforms DTAEA and KTDMOEA outperforms KTDMOEAv1. More details can be found from Tables 138-140 of the Suppmentary File. From this result, we can get the conclusion that, the proposed PS expansion/contraction method works better than the knowledge transfer in DTAEA, further confirming the effectiveness of our proposed knowledge transfer method.

\subsection{Performance Comparison on Other Changes in the NObj}
In the previous experiments, the NObj only increases or decreases one by one. This section aims to verify the performance of the proposed algorithm in the scenario where the NObj increases or decreases by more than one. Two different changing sequences where the NObj increases or decreases by one or two each time are designed as follows:
\begin{itemize}
\item The initial NObj is set as 2. Then, NObj firstly increases from 2 to 3. Then there are four changes with the first two changes increasing the NObj by two and then two changes decreasing the NObj by two. Lastly, the NObj decreases from 3 to 2 (simply denoted as `2-3-5-7-5-3-2').
\item The initial NObj is set as 7. Then, there are two changes where the NObj decreases by two. Later on, the NObj decreases from 3 to 2 and then increases from 2 to 3. In the last two changes, the NObj increases by two at each change (simply denoted as `7-5-3-2-3-5-7').
\end{itemize}

%

All experimental settings are set the same as in Section \ref{sec.Para} except for the frequency of change and the metric, which is set as 25 and HV, respectively, to save space. For Friedman and Nemenyi tests, the HV values that all algorithms get on one problem with one frequency of change at one independent run of 31 runs is regarded as an observation of the test. Therefore, there are 403 (13 problems and 1 frequency of changes and 31 independent runs) observed data.



\begin{figure}
\hspace{-.3in}
  \centering
  \subfigure[For sequence `2-3-5-7-5-3-2'.]{
    \label{fig:KTHV} 
    \includegraphics[width=0.5\linewidth]{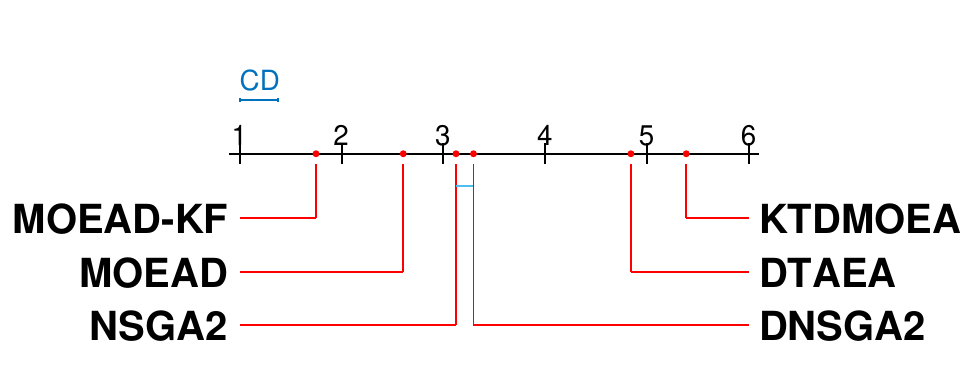}}
  \hspace{-.14in}
  \subfigure[For sequence `7-5-3-2-3-5-7'.]{
    \label{fig:KTGD} 
    \includegraphics[width=0.5\linewidth]{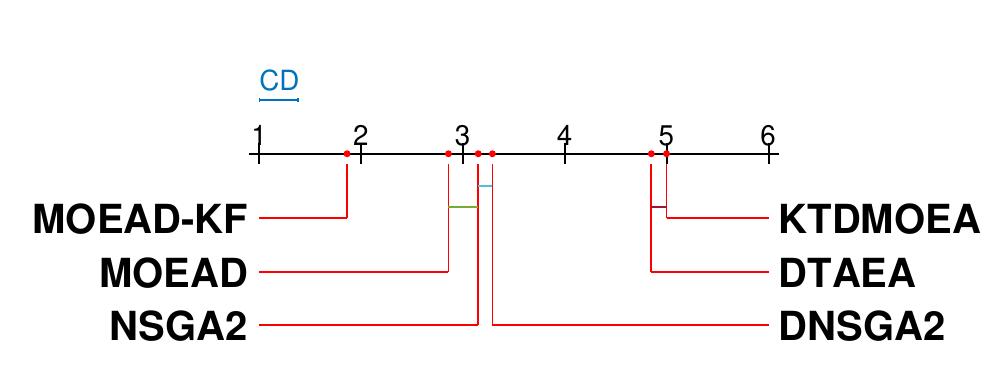}}
    \hspace{-.3in}
    \vspace{-0.3cm}
  \caption{Friedman ranking among HV of optimized solutions at the last generation by all algorithms in the changing sequences `2-3-5-7-5-3-2' and `7-5-3-2-3-5-7', respectively.}
  \label{fig:NObjChangesMoreThan1} 
  \vspace{-0.4cm}
\end{figure}

Figure \ref{fig:NObjChangesMoreThan1} (a) and (b) presents the Nemenyi post-tests results among HV values of optimized solutions at the last generation by all compared algorithms in two changing sequences `2-3-5-7-5-3-2' and `7-5-3-2-3-5-7', respectively. Friedman detects significant differences in average accuracy for HV with a p-value of 9.6123e-232 and 9.6955e-181, respectively for these two changing sequences.

Overall, it can be observed from Figure \ref{fig:NObjChangesMoreThan1} in the changing sequence of `2-3-5-7-5-3-2', our proposed KTDMOEA performs the best among all compared algorithms. In another changing sequence of `7-5-3-2-3-5-7', both KTDMOEA and DTAEA performs the best. More details can be found in Tables 141 and 142 in the Supplementary File.

\subsection{Impact of Algorithm Parameters}
\label{sec.ImpactAlgPara}
In the process of PS expansion, there is a parameter $\theta$ to set which is the number of solutions to generate along each expansion direction. In this section, different values of this parameter will be set to verify whether different parameter settings affect the performance of KTDMOEA.

All experimental settings are set the same as in Section \ref{sec.Para} except for the frequency of change and the metric, which is set as 25 and HV, respectively, to save space. The changing sequence is that the NObj firstly increases from 2 to 7 and then decreases from 7 to 2 one by one. There are three KTDMOEAs (denoted as KTDMOEA-1, KTDMOEA-2 and KTDMOEA-4), which has the value of $\theta$ as 1, 2 and 4, respectively. In order to verify whether different parameter settings affect the performance of KTDMOEA, the Friedman and Nemenyi tests on 5 state-of-the-arts and 3 KTDMOEAs are conducted to indicate the significant differences among them. The HV values that all algorithms get on one problem with one frequency of change at one independent run of 31 runs is regarded as an observation of the test. Therefore, there are 403 (13 problems and 1 frequency of changes and 31 independent runs) observed data.

\begin{figure}[h]
	\centering\vspace{-0.4cm}
	\includegraphics[width=0.25\textwidth]
	{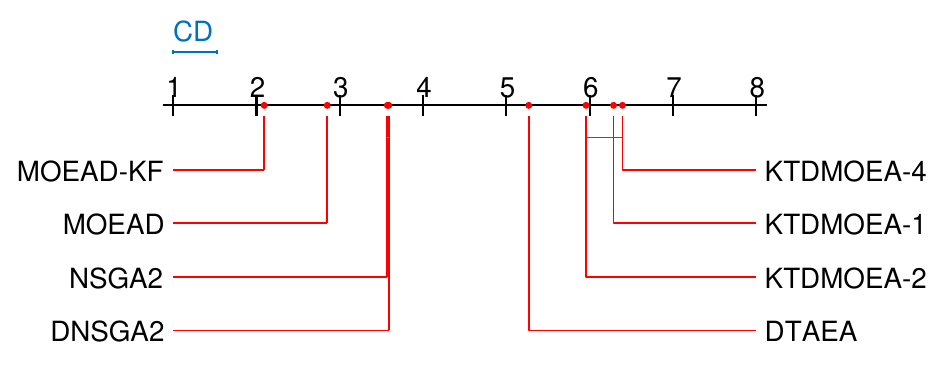}\vspace{-0.5cm}
\caption{Friedman ranking among HV of optimized solutions at the last generation by 6 state-of-the-arts and 3 KTDMOEAs with different values of parameters theta (1, 2 and 4) in the changing sequence of firstly increasing the NObj from 2 to 7 and then decreasing it from 7 to 2, both one by one.}
	\label{fig:KTDMOEA124}
	\vspace{-0.1cm}
\end{figure}

Figure \ref{fig:KTDMOEA124} presents the Nemenyi post-tests results among HV values of optimized solutions at the last generation by 8 algorithms in the changing sequence of firstly increasing the NObj from 2 to 7 and then decreasing it from 7 to 2, both one by one. Friedman detects significant differences in average accuracy for HV with a p-value of 1.7925e-281.

It is clear that the three KTDMOEAs get the best HV values among all algorithms. It can be found from Figure \ref{fig:KTDMOEA124} that there is no significant difference among the three KTDMOEAs with different setting of the parameter $\theta$. These results have verified that the performance of the proposed PS expansion/contraction is not sensitive to the setting of the parameter $\theta$. The performance of the proposed KTDMOEA against the existing algorithms is not affected by the setting of different parameter values.

\section{Conclusion}
\label{sec:conclusion}

It has been investigated in this paper that existing work cannot handle well DMOPs with a changing NObj and more complex PF shapes (convex, discontinuous and mixed shape of convex and concave) and fitness landscapes (nonseparability and deceptiveness). The main reason is the lack of sufficient population diversity right after dynamic changes. To solve this problem, two research questions are studied, first how to transfer knowledge so as to enhance diversity and second how the knowledge transfer helps optimization. In order to answer both research questions, inspired by the characteristic of DMOPs with a changing NObj, a dynamics handling strategy--PS expansion/contraction is proposed. As a result, a new algorithm, KTDMOEA, is designed to make use of this strategy. Experimental studies have demonstrated the effectiveness of the proposed knowledge transfer in enhancing the diversity right after changes and assisting the optimization under different number of objective changing sequences.

In comparison with the state-of-the-art in solving DMOPs with a changing NObj, our KTDMOEA achieved the best performance according to HV, GD and MS metrics across a number of test functions. We argue that is is important to use both DTLZ and WFG functions to build dynamic benchmarks because they provide rather different problem characteristics. KTDMOEA performed well on different problems under different parameter settings under different frequencies of change. 

As expected, no algorithm would be the best on all possible problems. According to the details in the Supplementary File, there are several problems on which KTDMOEA did not outperform existing algorithms. Although we have done initial analysis of these, as reported in the Supplementary File, more in-depth analysis will be our next work in the future. In addition, testing our proposed approach on real problems is one of our future work.

\section*{Acknowledgment}
This work has received funding from the European Union’s Horizon 2020 research and innovation programme under grant agreement number 766186. The work was also supported by the Guangdong Provincial Key Laboratory (Grant No. 2020B121201001), the Program for Guangdong Introducing Innovative and Enterpreneurial Teams (Grant No. 2017ZT07X386), and Shenzhen Science and Technology Program (Grant No. KQTD2016112514355531).


\end{document}